\documentclass[11pt,authoryear]{elsarticle}

\usepackage[margin=1in]{geometry}
\usepackage{amsmath}
\usepackage{amsfonts}
\usepackage{amssymb}
\usepackage{mathrsfs}
\usepackage[hidelinks]{hyperref}

\usepackage[american]{babel}

\usepackage{tipa}
\usepackage{graphicx}

\usepackage{float}
\usepackage{multirow}

\begin{document}

\begin{frontmatter}

\title{Local and non-local  dependency learning and emergence of rule-like representations in speech data by Deep Convolutional Generative Adversarial Networks}

\author[1]{Ga\v{s}per Begu\v{s}}
\ead{begus@berkeley.edu}
  \address[1]{Department of Linguistics, University of California, Berkeley, United States of America}
  
  \journal{Computer Speech \& Language}
  
\begin{abstract}

This paper argues that training Generative Adversarial Networks (GANs) on local and non-local dependencies in speech data offers insights into how deep neural networks discretize continuous data and how symbolic-like rule-based morphophonological processes emerge in a deep convolutional architecture. 
Acquisition of speech has recently been modeled as a dependency between latent space and data generated by GANs  in \citet{begus19}, who models learning of a simple local allophonic distribution. We extend this approach to test learning of local and non-local phonological processes that include approximations of morphological processes. We further parallel outputs of the model to results of a behavioral experiment where human subjects are trained on the data used for training the GAN network. Four main conclusions emerge: (i) the networks provide useful information for computational models of speech acquisition even if trained on a comparatively small dataset of an artificial grammar learning experiment; (ii) local processes are easier to learn than non-local processes, which matches both behavioral data in human subjects and typology in the world's languages. This paper also proposes (iii) how we can actively observe the network's progress in learning and explore the effect of training steps on learning representations by keeping latent space constant across different training steps. Finally, this paper shows that (iv) the network learns to encode the presence of a prefix with a single latent variable; by interpolating this variable, we can actively observe the operation of a non-local phonological process. The proposed technique for retrieving learning representations has general implications for our understanding of how  GANs discretize continuous speech data and suggests that rule-like generalizations in the training data are represented as an interaction between variables in the network's latent space.
\end{abstract}
\end{frontmatter}

\section{Introduction}
\label{intro}

The discussion between connectionist and symbolic approaches to language and human cognition in general has long been in the focus of computational cognitive science (\citealt{rumelhart86,mcclelland86,marcus01}, i.a.). Phonetic and phonological data are
uniquely appropriate for addressing this problem. Over a century-long tradition of scientific study of acoustic and perceptual phonetics (for an overview, see \citealt{allan13}) that deals with physical properties of speech sounds provides a solid understanding of the continuous data that hearing infants acquire language from: raw acoustic speech. Phonology is the study of how humans analyze, discretize, self-organize, and manipulate continuous speech data into discretized mental representations called \emph{phonemes}. The scientific study of phonology, too, has an over-a-century long history (for an overview, see \citealt{hulst13}), which resulted in a solid understanding of local and non-local discrete dependencies in human speech. Phonetic and phonological data and analysis are thus uniquely appropriate for probing what deep convolutional networks can and cannot learn, how discrete representations can emerge in deep neural networks, and how their performance can be paralleled to human behavior. Despite these advantages, the majority of neural network interpretability studies focus on non-linguistic visual data or syntactic/semantic levels, the latter of which lack a continuous component.

Computational models of speech acquisition have a long history. The majority of models, however, operate with abstract and already discretized data rather than raw acoustic inputs \citep{mccleland86,gaskell95,plaut99}. Deep neural network models of phonetic and phonological data operating with raw acoustic inputs emerged only recently. Several proposals model phonetic learning with deep autoencoder models \citep{rasanen16,alishahi17,eloff19,shain19,chung20}. Autoencoders learn to reduce data and encode data distributions in latent representations: they are trained on reproducing inputs by generating outputs from a reduced latent space. Inputs are thus directly connected to the outputs with an intermediate latent space that is reduced in dimensionality. Clustering analyses on the latent space show that the networks trained on phonetic data learn approximations of phonetic features based on phonetic similarity \citep{rasanen16,alishahi17,eloff19,shain19}.

While the reduced dimensionality in the autoencoder architecture approximates phonetic features based on phonetic similarity, the proposals do not model phonological processes. The human language learner has to acquire not only the identity of individual sounds based on acoustic similarity (as approximately modeled by the proposals using the autoencoder architecture), but also to manipulate those sounds in a given phonetic context. For example, a voiceless bilabial stop /p/ in English can surface as aspirated [\textipa{p\super h}] (produced with aspiration or a puff of air) before stressed vowels or as unaspirated [p] (without aspiration or a puff of air) if a fricative [s] precedes it. A minimal pair illustrating this distribution is \textipa{["\textbf{p\super h}It]} `pit' and \textipa{["s\textbf{p}It]} `spit'. The learner needs to learn not only to output voiceless bilabial stop, but also to shorten the aspiration time (VOT) when an [s] precedes it. Autoencoders are also trained on replicating output data as closely as possible to the input data, which is not desirable in models of language acquisition. While dimensionality reduction in autoencoders is unsupervised, input-output pairing is not.  

To model phonetic learning simultaneously with the learning of simple allophonic processes, \cite{begus19} proposes that speech acquisition can be modeled as a dependency between the latent space and generated data in the Generative Adversarial Networks.
Generative Adversarial Networks (GAN), first proposed by \citet{goodfellow14}, have not been used for modeling language acquisition, despite several advantages that this architecture features for computational models of language learning. GAN models are unsupervised and fully generative, which means that a deep convolutional network outputs innovative data that have no direct link to the training data (unlike, for example, in the autoencoder architecture). In other words, deep convolutional networks in the GAN architecture need to learn to output data from some random distribution. 

\citet{begus19} argues that deep convolutional networks in the GAN architecture encode discretized phonetic and phonological representations in the latent space. 
A computational experiment is conducted on a GAN implementation for audio  (as proposed in \citealt{donahue19} based on \citealt{radford15}) by training the networks on an phonologically local allophonic distribution in English, where voiceless stops surface as aspirated word-initially before a stressed vowel (e.g.~in [\textipa{"p\super hIt}] `pit'), except if a sibilant [s] precedes the stop  (e.g.~in [\textipa{"spIt}] `spit'). The network learns the allophonic distribution and encodes phonetically and phonologically meaningful features in its latent space. 

Based on this  local allophonic distribution, \cite{begus19} proposes a technique for identifying and manipulating variables in the latent space in the GAN architecture that correspond to desired phonetic and phonological representations. \cite{begus19} argues that the network uses a subset of latent variables to encode presence of a sound in the output (e.g.~[s]). By manipulating the identified variables, especially well beyond the training range (as proposed in \citealt{begus19}), we can actively force the sound in and out of the generated outputs. Moreover, a linear interpolation of the chosen latent variables from marginal values results in almost linear reduction of the amplitude of the frication noise of [s] --- a linguistically meaningful unit \citep{begus19}.

The goal of this paper is to argue that using the technique proposed in \cite{begus19}, we can model not only simple allophonic processes, such as English deaspiration, but also local and non-local phonological processes that are based on what would be approximated as morphology (morphophonological alternations) that resemble rule-like behavior. We also argue that we can parallel human behavioral experiments with performance of the deep convolutional networks that are trained on the same data as used in behavioral experiments.  In general, natural languages strongly prefer local over non-local processes, both in phonology and on other levels such as morphology and syntax \citep{finley11,finley12,mcmullin19,white18}. In fact, the vast majority of phonological processes in the world's languages are local (targeting adjacent sounds) \citep{finley11}, with only a few processes, such as harmony, operating on non-adjacent sounds. 
Behavioral experiments show that local processes are easier to learn than non-local processes \citep{finley11,finley12,mcmullin19,white18}.  In this paper, we test the learning of local and non-local phonological dependencies, and show that local processes (such as postnasal or intervocalic devoicing) are easier to learn for the networks than non-local vowel harmony. We parallel success rates in the computational model to behavioral data --- an artificial grammar learning experiment in which human subjects are trained on the same data (Section \ref{paralleling}). This type of combining artificial grammar learning experiments and computational models has the potential to reveal similarities in learning biases between human subjects and deep convolutional networks, and shed light on how domain-general learning biases that require no language-specific mechanisms can result in the typological prevalence of local processes and the rarity of non-local processes. 

Specifically, we test the learning of non-local vowel harmony and several local devoicing patterns. Vowel harmony is a phonological process, usually non-local, in which a vowel becomes more similar to another vowel in a word. For example,  the plural morpheme in Turkish surfaces as [\textipa{lAr}] after root vowels that are back and as [\textipa{ler}] if the root vowel is front \citep{kabak11}: [\textipa{d\textbf{A}l-l\textbf{A}r}] `branches' and [\textipa{j\textbf{e}r-l\textbf{e}r}] `places' \citep{kabak11}.

 In formal phonological analysis, phonological computation is formalized with rewrite rules that operate as symbolic feature manipulation \citep{ch68}. As argued by \cite{marcus99} and several other works (\citealt{ch68,heinz10,berent13}, i.a.), ``algebraic rules'' are required to derive a set of surface outputs such as Turkish [\textipa{d\textbf{A}l-l\textbf{A}r}] and  [\textipa{j\textbf{e}r-l\textbf{e}r}] from stored inputs. The stored mental representation of the prefix can be posited as /\textipa{lAr}/. The role of phonological grammar is to derive the two surface forms (outputs) from the stored mental representation (input).

 Sounds are represented with matrices of binary features that distinguish meaning (e.g.~[+syllabic, + front] means a front vowel). Vowel harmony can be formalized with a simple rewrite rule (in \ref{formalrule}) that identifies vowels ([+\text{syllabic}]) and assigns the same value ($\alpha$) of feature [$\pm$front] as in the vowel that follows it (interrupted by any number of consonants C$_0$). The formalism is illustrated in (\ref{formalrule}).

\begin{equation} [+ \text{syllabic}] \rightarrow [\alpha\ \text{front}] / \rule{2em}{.5pt}\text{C}_0[\alpha\ \text{front}] \label{formalrule}
\end{equation}

The discussion of symbolic representation vs.~connectionism has a long tradition in phonology. An influential proposal called Optimality Theory models phonology as an input-output pairing rather than a rule-based symbolic representation \citep{prince9304,legendre90}. Optimality Theory was directly influenced by earlier work on connectionism. Vowel harmony within this framework is modeled with the Agreement-by-correspondence proposal \citep{hansson10,rose04}: two sounds (such as the two vowels \textipa{[A]} in  Turkish [\textipa{d\textbf{A}l-l\textbf{A}r}]) are in correspondence and  share features, which, through surface optimization in the grammar, results in a harmonious process. Several independent facts support the approach of input-output optimization in phonology. However, both Optimality Theory and other proposals in phonology using neural networks \citep{mccleland86,gaskell95,plaut99} model local and non-local phonology with pre-assumed levels of abstraction, meaning that learning is not modeled from raw acoustic data but is already pre-discretized or requires language-specific mechanisms.

We argue that approximates to rule-based behavior emerge in deep convolutional networks even without any pre-assumed levels of abstraction (the networks are trained on raw acoustic inputs) and when models contain no language-specific parameters. The network discretizes the representation of a prefix in the output and uses only one latent variable (out of 100) to encode the presence of the prefix. Equivalents to non-local phonological rules emerge from an interaction between the variable that represents the prefix and a variable that generates some desired phonological process. We also argue that the same data used for training in the GAN architecture can be used to test phonological learning in artificial grammar learning experiments in human subjects. In fact, the paper argues that training GANs on relatively few data points yields, somewhat surprisingly, highly informative results (Section \ref{small}). This observation should open numerous opportunities for paralleling performance in deep neural networks and behavioral outcomes of artificial grammar learning experiments with human subjects. Finally, we outline a procedure to observe how the network learns dependencies as the training progresses and claim that the generator’s search through the space of phone-level combinations are linguistically interpretable (Section \ref{progression}).

\section{Materials}
\label{materials}

\subsection{Model}
\label{model}
The main characteristic of Generative Adversarial Network architecture \citep{goodfellow14}, and more specifically the DCGAN proposal by \cite{radford15}, are two deep convolutional neural networks that are trained in a minimax setting. The Discriminator learns to estimate realness of the data and minimize its own error rate \citep{brownlee19}. The Generator network learns to output data from a set of latent variables and maximize the Discriminator network's error. Initially, the Generator network produces noise, but as training progresses it becomes increasingly more successful in outputting data such that the Discriminator becomes less successful in distinguishing actual from generated data. 

The majority of GANs are trained on two-dimensional visual data; a shift to apply the architecture to the audio domain has occurred only recently with the work of \citet{donahue19} (WaveGAN). The model in \citet{donahue19}, used for training here, is based on the DCGAN architecture \citep{radford15} and features most of the same hyperparameters. The two main differences are that the Generator involves an additional layer and generates a one-dimensional output that corresponds to approximately 1 second of audio. The cost function is taken from the Wasserstein GAN proposal with gradient penalty (WGAN-GP) (as proposed in \citealt{arjovsky17} and \citealt{gulrajani17}). For all specifications of the model, see \citet{donahue19}.

\begin{figure}
\centering
\includegraphics[width=.75\textwidth]{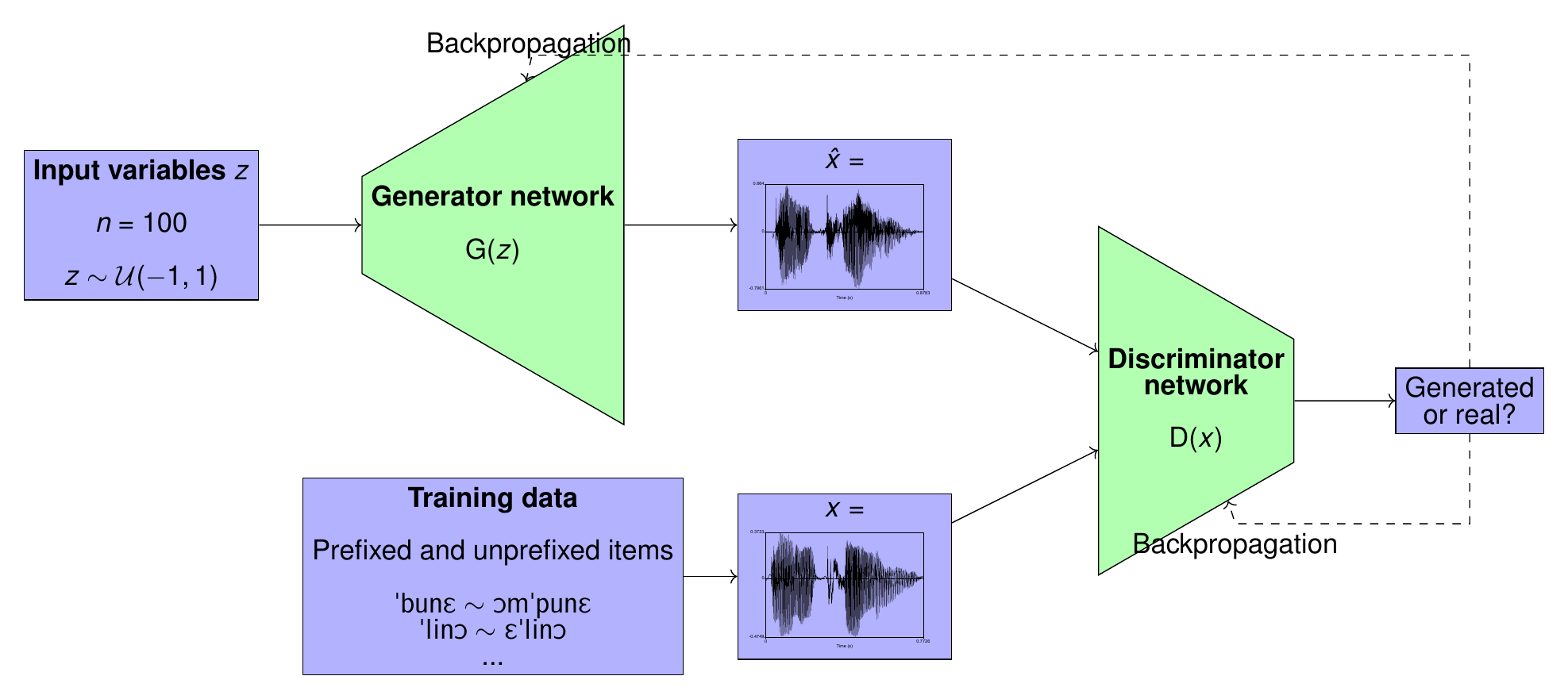}
\caption{\label{tikzflowMGAN} The GAN architecture schematized from \cite{goodfellow14,radford15,donahue19}  used in this paper with training data as described in Section \ref{data}.} 
\end{figure}

\cite{begus19}  proposes a technique for exploring learning representations in deep convolutional networks. For example, the network is trained on \#T\textsuperscript{h}V and \#sTV sequences from TIMIT (e.g.~[\textipa{p\super h\ae}] and [\textipa{sp\ae}]) and learns the conditional distribution: it mostly outputs short VOT (no aspiration) if an [s] precedes the stop and long VOT (aspiration) if no [s] precedes it. However, the Generator's outputs are not simply replications of its input: in about 12\% of outputs, the stop after  an [s] is aspirated ([\textipa{sp\super h\ae}]) and the VOT duration can be longer than in any \#sTV sequence in the training data. Additionally, the network occasionally outputs innovative sequences that lack a stop (e.g.~\#sV) or concatenate two stops (e.g.~\#TTV).   In other words,  the Generator learns the conditional allophonic distribution, but imperfectly so \citep{begus19}. The outputs with long VOT (aspiration) in the [s]-condition parallel stages in language acquisition: language-acquiring children also occasionally output stops with long VOT (aspiration) in the [s]-condition \citep{bond80}. 

In addition to observing learning in the GAN architecture with surface forms, we can identify individual latent variables that correspond to phonetic and phonological representations. \cite{begus19} proposes a technique for identification of the variables by regressing the annotated outputs to the randomly sampled latent space. Predictions of several regression models are tested  in \cite{begus19} to avoid assumptions of linearity: generalized additive models with various shrinkage techniques, linear logistic regression, Lasso logistic regression, and random forest models. The technique identifies latent variables ($z$; see  Figure \ref{tikzflowMGAN}) that correspond to presence of [s] in the output. Moreover, it is shown that the relationship between the individual latent variables (e.g.~those identified as representing [s]) and the presence of [s] in. the generated data are often linear, even when non-linear regression is used for testing. 

Given this linear relationship, we can identify variables that correspond to a desired phonetic property and identify whether the property correlates with positive or negative values of the variable. Individual $z$-variables are uniformly distributed during the training with the interval $(-1, 1)$. When set to a value identified as corresponding to presence of a desired phonetic feature, the output contains a significantly higher proportion of this property. Crucially, \cite{begus19} shows that manipulating the identified variables beyond the values in the training range $(-1, 1)$, such as to $\pm4.5$, results in an increased presence and amplitude of the desired phonetic representation. In other words, as we interpolate a variable identified as representing an [s] in the output, the amplitude of [s] increases or decreases. We can thus actively force a phonetic or phonological feature in the output. That the proposed technique indeed identifies variables corresponding to the presence of [s] is suggested by an independent generative test in \cite{begus19}. While explorations of latent space and representation learning in GANs have been conducted before on visual data \citep{radford15}, the proposals, to the author's knowledge, do not use single variables to explore their meaningful equivalents in the output and do not utilize interpolation to extreme values beyond the training range.\footnote{\cite{radford15} uses averaging over $z$-variables in some cases and performs logistic regression on the second to last convolutional layer.}

 \cite{begus19} thus argues that the Generator network learns a local allophonic distribution as well as learns to encode phonetic and phonological representations with a subset of variables in the latent space. 
While the Generator network represents  [s] in the latent space with a subset of variables in \cite{begus19}, the cutoff between variables associated with presence of [s] and the rest of the latent space is not completely categorical. The Generator network does not associate the presence of [s] with a single variable: seven $z$-variables are associated with the representation of [s]. There is a notable cutoff between the regression estimates of the seven highest variables and the rest of the latent space, but the difference is not substantial or categorical. Training data in \cite{begus19} is sliced from TIMIT \citep{timit}, which is considerably more variable than the training data in this experiment. As is argued in Section \ref{latentspace}, discretization of some morphophonological representation (e.g.~presence of the prefix) is substantial in the current experiment. It appears that less variable data results in a more rapid discretization.

\subsection{\label{data}Data}

The training data (from \citealt{begusCatalysis}) contain evidence for one non-local phonological process --- vowel harmony --- and four local processes: (i) post-nasal devoicing of stops ([\textipa{"\textbf{b}Alu}] $\sim$ [\textipa{Om"\textbf{p\super h}Alu}]), (ii) post-nasal occlusion with devoicing of voiced fricatives ([\textipa{"\textbf{v}i\*r@}] $\sim$ [\textipa{Em"\textbf{p\super h}i\*r@}]), (iii) intervocalic devoicing of stops ([\textipa{"\textbf{b}ulO}] $\sim$ [\textipa{O"\textbf{p}\super hulO}]), and (iv) intervocalic fricativization with devoicing of stops ([\textipa{"\textbf{b}O\*r@}] $\sim$ [\textipa{O"\textbf{f}O\*r@}]). These processes are triggered by prefixes; the training data thus contain bare (unprefixed) and prefixed forms of lexical items of the shape (\textsc{prefix-})CVCV and (\textsc{prefix-})CVC (C = consonant, V = vowel), e.g.~[\textipa{"\*rinu}] $\sim$	[\textipa{En"\*rinu}]. The items are all nonce words in English, so that the same dataset can be used in the behavioral experiment with human subjects (Section \ref{paralleling}).

\subsubsection{Non-local processes}

Non-local vowel harmony is triggered by the first vowel of the base (unprefixed) form and results in two different vowel qualities of the prefix, [\textipa{E}] and [\textipa{O}]. The descriptive generalization is the following: the vowel of the prefix is [\textipa{E}] if the first vowel of the lexical item is [\textipa{E, i}] and  [\textipa{O}] if the vowel is [\textipa{A, O, u}].  For example, a lexical item such as [\textipa{"linO}] has a prefixed form  [\textipa{En"linO}] with a front vowel in the prefix [\textipa{En-}] because the first vowel in the lexical item [i] is front. A lexical item such as [\textipa{"luru}] has a prefixed form with [\textipa{On-}]:  [\textipa{On"luru}] because the first vowel of the lexical item [u] is not front. 
The experiment thus features a similar case of vowel harmony as the Turkish example (see Section \ref{intro}).

The computational experiment presented here tests the learning of non-local vowel harmony. That the process tested here is phonologically non-local is clear from Table \ref{examples}: the sounds in correspondence (the vowel of the prefix and the first vowel of the lexical item) are always separated by one or two consonants.

\subsubsection{Local processes}

In addition to non-local vowel harmony, the training data contain evidence for four local processes that are triggered by the prefix.  Two processes are triggered by a nasal sound in the prefix VN-. 16 unprefixed-prefixed pairs (32 items total) contain evidence for post-nasal devoicing (D $\rightarrow$ T / N\rule{2em}{.5pt}), where a voiced stop devoices if a nasal precedes it: [\textipa{"\textbf{b}Alu}] $\sim$ [\textipa{Om"\textbf{p\super h}Alu}]. In another 16 pairs (32 items total), a voiced fricative gets devoiced and occluded when a nasal precedes it (Z $\rightarrow$ T / N\rule{2em}{.5pt}): [\textipa{"\textbf{v}i\*r@}] $\sim$ [\textipa{Em"\textbf{p\super h}i\*r@}]. The other two processes are triggered by the V-prefix. The evidence for intervocalic devoicing, where voiced stops devoice intervocalically (D $\rightarrow$ T / V\rule{2em}{.5pt}V) is present in 16 unprefixed-prefixed pairs (32 items total), e.g.~[\textipa{"\textbf{b}ulO}] $\sim$ [\textipa{O"\textbf{p}\super hulO}]. Another 16 pairs (32 items total)  contain evidence for intervocalic fricativization and devoicing, where voiced stops fricativize and devoice (D $\rightarrow$ S / V\rule{2em}{.5pt}V) between vowels (triggered by the prefix), e.g.~[\textipa{"\textbf{b}O\*r@}] $\sim$ [\textipa{O"\textbf{f}O\*r@}].
In the 54 remaining pairs (108 total), no consonantal changes are present, e.g.~[\textipa{"jAlu}] $\sim$ [\textipa{O"jAlu}] or [\textipa{"\*rinu}]	$\sim$	[\textipa{En"\*rinu}]. 

Because the learning of non-local processes is predicted to be more difficult than that of local processes, the training data contain substantially more evidence for the non-local process. All items in which C$_1$ is constant as well as those in which it changes contain evidence for the  non-local vowel harmony process.  Of 270 training items, there are 117 unprefixed items with 117 corresponding prefixed forms, all of which contain evidence for vowel harmony (234 total). The remaining items (36) only include unprefixed forms (for testing learning).
There is thus a substantial difference in the amount of training data that contain evidence for the non-local process (117 pairs, 234 altogether) and the four local processes (16 pairs each). Even if all four local processes are pooled together, the data still contain only 64 pairs containing evidence for the four local processes (128 altogether). Table \ref{examples} illustrates the training data: each slot is filled with a transcribed example from the training data. The entire training in IPA transcription is given in Appendix Tables  \ref{nonce1}, \ref{nonce1a}, \ref{nonce2}, \ref{nonce2a}, \ref{nonce3}, \ref{nonce3a}, and \ref{test3a}.

In addition to the local and non-local processes described above, the data contain evidence for a local assimilation process which is somewhat less relevant to our experiment: if the prefix contains a nasal stop (VN-), the place of articulation of the nasal stop depends on the first consonant of the root (C$_1$). The nasal  surfaces as labial [m] before the labials ([p] and [f]), and as an alveolar [n] elsewhere. Spectral differences are minimal between the two conditions,  which is why a detailed analysis of this process is not possible in the computational experiment; the main purpose for including this assimilation in the data is for the behavioral experiment to include an English-like process (to not raise the attention of the subjects) and to facilitate the reading task for the speaker who recorded the stimuli.

The computational experiment tests the learning of the local devoicing processes and non-local vowel harmony that target the \textsc{prefix} (VN- or V-). In order to control for the potential effects of other segments on the learning of the targeted processes, we balance the experimental design as much as possible. The number of lexical items with the front vowel in V$_2$ is, in all but three pairs, equivalent for every C$_1$ condition.  In other words, if there are four [d]-initial items that devoice and have frontness harmony (V$_2$ is front), there are also four items with backness harmony (V$_2$ is not front) for this condition.\footnote{There are two missing frontness harmony pairs in the non-changing \textipa{[p\super h]}- and \textipa{[t\super h]}-initial condition  and one missing backness harmony pair in the non-changing [l]-initial condition for the VN- prefix.} We also aim to balance the identity of C$_3$ and V$_4$  as much as possible, but balancing these positions is limited by the requirement that the items not be real words of English or too similar to real words (due to the artificial grammar learning experiment). Only [m, n, l, \textipa{\*r}, s] can be members of C$_3$, and these along with V$_4$ are relatively well balanced across the groups with changing C$_1$ (e.g.~approximately equal number of the same consonants across voiced-initial items that devoice and those that undergo devoicing with fricativization or occlusion), but not across other groups. A fully balanced design is difficult to achieve due to different groups and the nonce-word requirement, but given the relatively well balanced design, we do not expect undesired dependencies to affect the learning distributions of interest. 

The 270 items described above were presented in a simplified transcription (see Appendix Tables  \ref{nonce1}, \ref{nonce1a}, \ref{nonce2}, \ref{nonce2a}, \ref{nonce3}, \ref{nonce3a}, and \ref{test3a}) and read by a single female speaker of American English (see also \citealt{begusCatalysis}). The words were of the shape C$_1$V$_2$C$_3$,  C$_1$V$_2$C$_3$V$_4$,  \textsc{prefix}-C$_1$V$_2$C$_3$, and  \textsc{prefix}-C$_1$V$_2$C$_3$V$_4$. The prefixes were of the shape VN- and V-:  [\textipa{E}n-], [\textipa{O}n-], [\textipa{E}m-], [\textipa{O}m-], [\textipa{E}-],  and [\textipa{O}-]. The speaker was unaware of the exact objectives and details of the study and was compensated for her work. Recordings of training data were made in a sound-attenuated booth using a USBPre 2 (Sound Devices) pre-amp and Shure 53 Beta omnidirectional condenser head-mounted microphone in Audacity (originally sampled at 44.1 kHz and then downsampled to 16 kHz). 

\begin{table}
\centering
\caption{\label{examples}Examples of words used in training in the IPA transcription.}
\scalebox{.8}{\begin{tabular}{lllrr|rr|rrr}

  \hline\hline

\textbf{Prefix}&&&\multicolumn{2}{c}{\textbf{Labial}}&\multicolumn{2}{c}{\textbf{Coronal}}&\textbf{[j]}&\textbf{[l]}&    \textbf{[\textipa{\*r}]}\\\hline

\multirow{8}{*}{\textbf{VN-}}&\multirow{4}{*}{\textbf{C$_1$ constant}}&\multirow{2}{*}{\textbf{\textipa{E}-harmony}}&\textipa{"p\super himi}   &\textipa{"fim@}&  \textipa{"t\super hElO}  &\textipa{"sEnO}   &\textipa{"jim}&\textipa{"lEn}  &  \textipa{"\*rinu} \\    

&&&\textipa{Em"p\super himi} &\textipa{Em"fim@}&\textipa{En"t\super hElO}&\textipa{En"sEnO} &\textipa{En"jim}&\textipa{En"lEn}&\textipa{En"\*rinu}  \\ 

\cline{3-10}

&&\multirow{2}{*}{\textbf{\textipa{O}-harmony}}&\textipa{"p\super hO\*rO}   &\textipa{"fu\*r@}  &\textipa{"t\super hA\*ru}&  \textipa{"sAnu}  &\textipa{"jAlu}&\textipa{"lO\*r}&\textipa{"\*rOlO}\\

&&&\textipa{Om"p\super hO\*rO} &\textipa{Om"fu\*r@}&\textipa{On"t\super hA\*ru}&\textipa{On"sAnu}&\textipa{On"jAlu}&\textipa{On"lO\*r}&\textipa{On"\*rOlO}\\

\cline{2-10}

&\multirow{4}{*}{\textbf{C$_1$ changes}}&\multirow{2}{*}{\textbf{\textipa{E}-harmony}}&\textipa{"bE\*r@}&\textipa{"vir@}&\textipa{"dElO}&\textipa{"zi\*r@}&|&|&|\\

&&&\textipa{Em"p\super hE\*r@}&\textipa{Em"p\super hir@}&\textipa{En"t\super hElO}&\textipa{En"t\super hi\*r@}&|&|&|\\

\cline{3-10}

&&\multirow{2}{*}{\textbf{\textipa{O}-harmony}}&\textipa{"bAlu}&\textipa{"vOn@}&\textipa{"dun@}&\textipa{"zOlE}&|&|&|\\
&&&\textipa{Om"p\super hAlu}&\textipa{Om"p\super hOn@}&\textipa{On"t\super hun@}&\textipa{On"t\super hOlE}&|&|&|\\

\hline\hline

\multirow{8}{*}{\textbf{V-}}&\multirow{4}{*}{\textbf{C$_1$ constant}}&\multirow{2}{*}{\textbf{\textipa{E}-harmony}}&\textipa{"p\super hin@}&\textipa{"fini}&\textipa{"t\super hElO}&\textipa{"sEnO}&\textipa{"jim}&\textipa{"linO}	&\textipa{"\*rEl}	\\    

&&&\textipa{E"p\super hin@}&\textipa{E"fini}&\textipa{E"t\super hElO}&\textipa{E"sEnO}&\textipa{E"jim}&	\textipa{E"linO}&	\textipa{E"\*rEl}	\\

\cline{3-10}

&&\multirow{2}{*}{\textbf{\textipa{O}-harmony}}&\textipa{"p\super hO\*mO}	&\textipa{"fu\*r@}&\textipa{"t\super hOmO}&\textipa{"sAnu}&\textipa{"jAm}&\textipa{"lu\*ru}	&\textipa{"\*rAs}	\\

&&&\textipa{O"p\super hO\*mO}&\textipa{O"fu\*r@}&\textipa{O"t\super hOmO}&\textipa{O"sAnu}&\textipa{O"jAm}&	\textipa{O"lu\*ru}&\textipa{O"\*rAs}\\

\cline{2-10}

&\multirow{4}{*}{\textbf{C$_1$ changes}}&\multirow{2}{*}{\textbf{\textipa{E}-harmony}}&\textipa{"bEl@}&\textipa{"bEm@}&\textipa{"dEni}&\textipa{"dEmE}	&|&|&|\\

&&&\textipa{E"p\super hEl@}&\textipa{E"fEm@}&\textipa{E"t\super hEni}&\textipa{E"sEmE}&|&|&|\\

\cline{3-10}

&&\multirow{2}{*}{\textbf{\textipa{O}-harmony}}&\textipa{"bulO}&\textipa{"bO\*r@}&\textipa{"dA\*ru}&\textipa{"dAl@}&|&|&|\\
&&&\textipa{O"p\super hulO}&\textipa{O"fO\*r@}&\textipa{O"t\super hA\*ru}&\textipa{O"sAl@}&|&|&|\\

\hline\hline

\end{tabular}}
\end{table}

The data in the form of sliced audio files for each item (approximately 1 s long padded with silence)  is fed to the model randomly in mini-batches of 64. The bare unprefixed and prefixed forms are not paired in any way during training.

\section{Results}

One advantage of the GAN architecture is that the Generator network outputs innovative data that are linguistically interpretable \citep{begus19}. Innovative outputs are often sporadic and do not allow for a full quantitative analysis, which nonetheless does not make them less informative. It is important to describe innovative outputs and how they can inform us about the learning of speech data in deep convolutional networks. 
In Sections \ref{small} and \ref{progression} we present results from an exploratory study of the network's innovative outputs based on an acoustic analysis of spectra. In Sections \ref{latentspace}, \ref{lnlp}, and \ref{erbb} we present a quantitative analysis of the generated outputs.\footnote{Generated data and trained models are available at \url{https://doi.org/10.17605/OSF.IO/A9WMY}.}

\subsection{Small data sets}
\label{small}
The total unique data points (audio recordings of the words with the structure described in Section \ref{data}) that the network is trained on is 270. Despite the small amount of training data, the model generates outputs that closely resemble human speech, are interpretable, analyzable, and highly informative. This stands in contrast to some recent studies of neural network models on the syntactic level that require very large training datasets and do not improve substantially with more data \citep{schijndel19}. As is argued below, the GANs do not overfit, but produce innovative data that are linguistically interpretable despite the small training data set. This finding should open up numerous possibilities for further exploration of learning representations in deep convolutional networks: it is generally assumed that GANs and deep convolutional networks require large amounts of data, which could be prohibitive for research questions that require smaller training datasets.

We analyze outputs of the Generator network at four training steps: after 7453 ($\sim$ 8833 epochs), 9740 ($\sim$ 11543 epochs), 14900 ($\sim$ 17659 epochs), and 20990 ($\sim$ 24877 epochs) steps. The number of steps chosen is based on maximizing clarity of the acoustic outputs that need to be appropriate for acoustic analysis and minimizing the number of steps used for training (for guidelines, see \citealt{begus19}).

Some generated outputs are phonetically very similar to the input equivalents, as illustrated in Appendix A Figure \ref{dino7453ACL2020}. The network, however, also generates outputs that substantially violate the input data. The Generator network trained after 7453 steps, for example, outputs a sequence that can be transcribed as [\textipa{"dinO}], yet the training data lacks this sequence altogether. The closest neighbor to the innovative  [\textipa{"dinO}] in the training data is  [\textipa{"dEnO}] (see Figure \ref{dino7453ACL2020}). There are numerous other such generated outputs that violate the training data, but are linguistically valid and interpretable. For example, 23.2\% of outputs violate the training data with respect to vowel harmony (see Section \ref{lnlp}).

To further quantify the proportion of innovative outputs that are linguistically interpretable, we transcribe 200 randomly generated outputs from a network trained after 20990 steps. The phonemic structure is impossible to determine in only 13 of the 200 outputs (6.5\%). In the majority of these 13 outputs, the generated audio resembles speech and includes periodic vibration, but spectrogram structure is too noisy for identification of clear phonetic structure for parts of the output or the entire output.  On the other hand, in the majority of cases (187 or 93.5\%), the generated outputs have a clear and identifiable phonetic structure. Moreover, the Generator clearly learns the structural phonotactic properties of the input data. In all outputs with an identifiable structure, the network outputs items with the structure CVCV,  CVC, \textsc{prefix}-CVCV, or  \textsc{prefix}-CVC. The network also learns more specific distributional patterns. For example, training data lacks nasal consonant in the initial position (C$_1$ in C$_1$V$_2$C$_3$V$_4$ or C$_1$V$_2$C$_3$ is never a nasal, but either an obstruent or [l, \textipa{\*r}, j]; see also Tables \ref{dino7453ACL2020}, \ref{nonce1}, \ref{nonce1a}, \ref{nonce2}, \ref{nonce2a}, \ref{nonce3}, \ref{nonce3a}, \ref{test3a} in Appendix). On the other hand, C$_3$ never features an obstruent with the exception of [s] ([\textipa{p\super h, t\super h, b, d, f, v, z}]) in the training data. Finally, obstruents are always voiceless in prefixed forms (e.g.~[\textipa{En"\textbf{t\super h}ilO}] for [\textipa{"dilO}]).  All 187 outputs conform to all these distributional patterns. 

 Crucially, the Generator does not simply replicate inputs. While all 187 outputs conform to the global distributional patterns of the training data, 78/187 are unique combinations of sequences that are absent from the training data. 15 out of these 78 outputs are disharmonious cases. Yet, even if they are taken out of consideration, the generator outputs  63/187 (33.7\%) of outputs that conform to distributional and phonotactic patterns of input data, but feature unique phoneme sequences that are absent from the training data. For example, the network outputs [\textipa{"bO\*rO}], [\textipa{O"t\super hOn@}], and [\textipa{"t\super hini}] which conform to the phonotactic patterns of the training data, but are not present in precisely these particular combinations of segments in the training data.

Innovative outputs that violate training data distributions in linguistically interpretable ways constitute strong evidence against overfitting in the GAN architecture: even with very small datasets and a relatively high number of epochs, the Generator does not overfit. This is in line with previous evidence that GANs generally do not overfit \citep{adlam19,donahue19}, but here we additionally argue that GANs don't overfit even with small training datasets (N = 270).

\subsection{Progression of learning}
\label{progression}

One advantage of the exploratory study of GANs outputs is that we can follow how dependencies in speech are learned by the network at different training steps. We propose that the progression of learning can be observed by keeping the latent space constant and generating data at different training stages of the Generator network. 
This provides crucial information on how the number of training steps influences the Generator's outputs and learning representations --- an area that is relatively understudied. Testing the effect of training steps on learning representations using speech data should reveal further insights into neural network interpretability, as is argued below.

\begin{figure}
\centering
\includegraphics[height=.18\textheight]{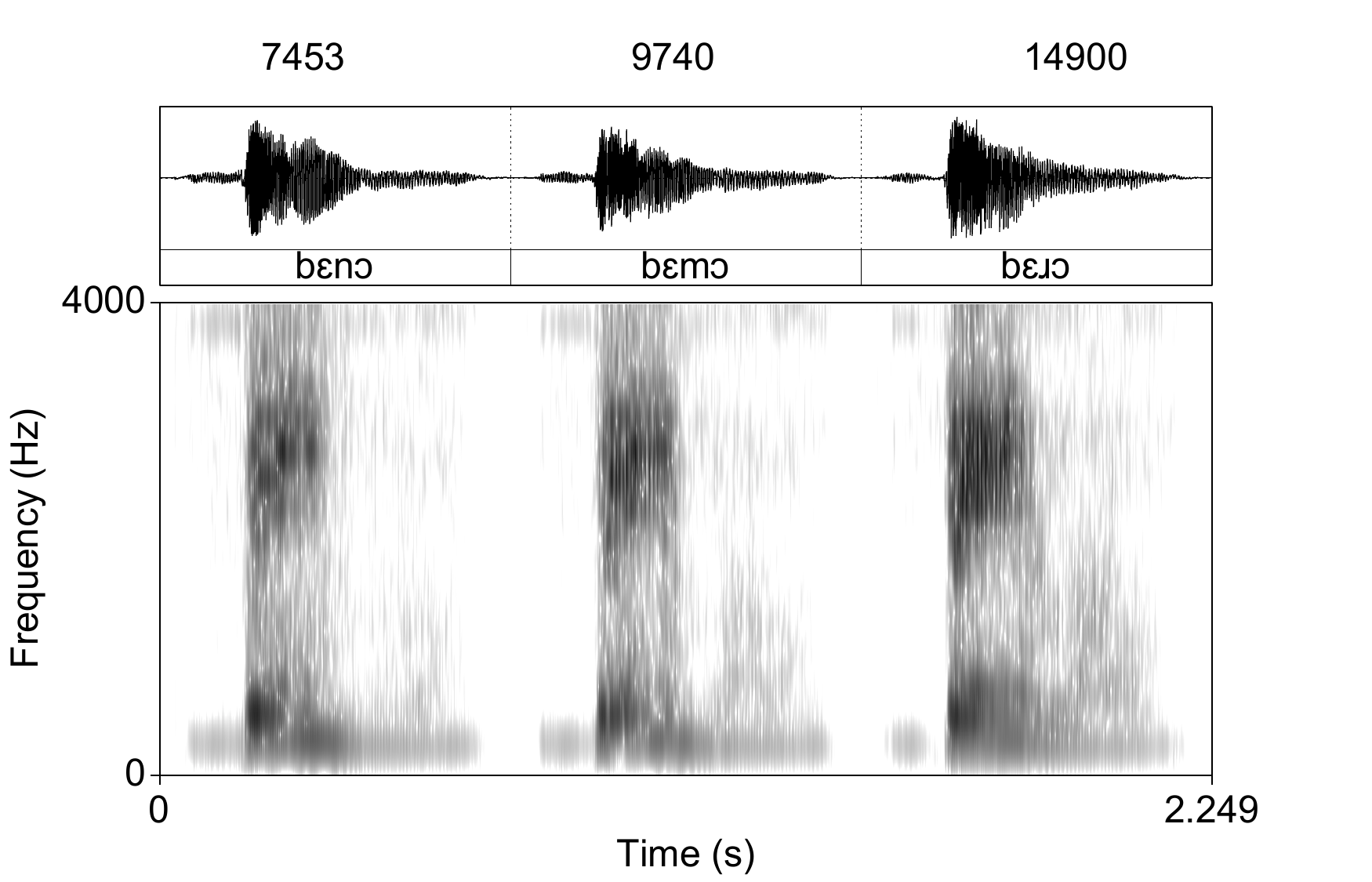}
\includegraphics[height=.18\textheight]{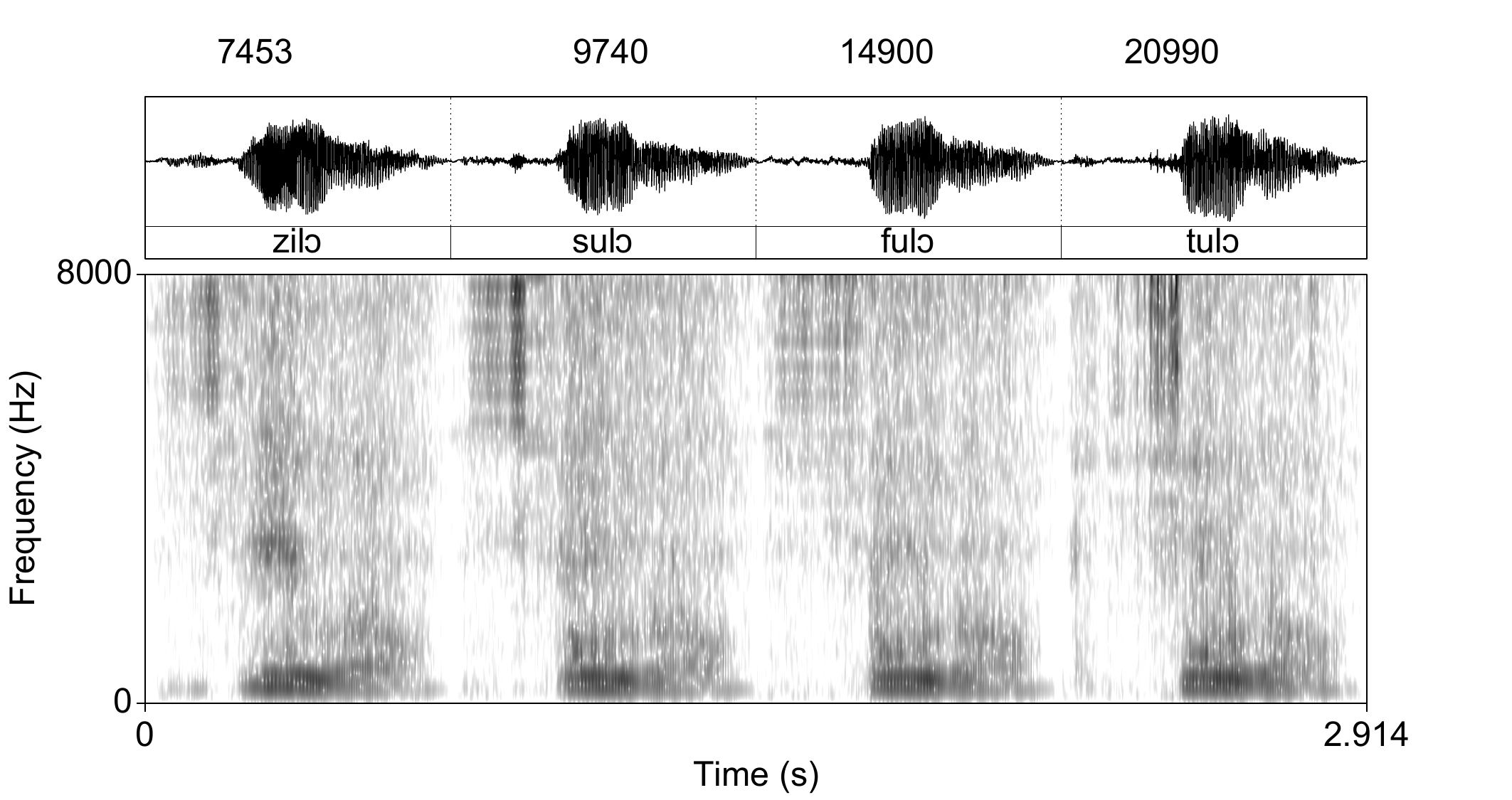}
\caption{\label{benoProgressACL2020} \textbf{(left)} Waveforms and spectrograms (0--4000 Hz) of three generated  samples with the same values of latent variables at three different training steps. \textbf{(right)} Waveforms and spectrograms  (0--8000 Hz) of four generated  samples with the same values of latent variables at four training steps showing devoicing, change of place of articulation, and occlusion.} 
\end{figure}

We propose that by analyzing generated outputs at different training steps with latent space kept constant, we can actively follow how the network corrects the outputs that violate distributions in the data. For example, at 7453 steps, the network generates an innovative output that violates the training data: [\textipa{"bEnO}]. At 9740 training steps, the network outputs [\textipa{"bEmO}] for the same latent space variables. This output still violates the data: none of the words in the training data was of the exact shape [\textipa{"bEmO}]. At 14900 steps, the network outputs [\textipa{"bE\*rO}] (for the same latent space), which corresponds to [\textipa{"bE\*rO}] in the training data (Figure \ref{benoProgressACL2020}).\footnote{The Generator outputs only waveforms; spectrograms are provided for the purpose of acoustic analysis.} 

In a related example, the proposed method allows us to follow how the network searches through the space of possible segment combinations using linguistically valid strategies. Figure \ref{benoProgressACL2020} shows an output  [\textipa{"zilO}] for which there is no direct equivalent in the training data. The spectrogram shows a clear voicing bar and frication noise in the high frequencies, characteristic of a [z]. At 9740 steps, the network devoices the initial consonant C$_1$, but keeps its frication noise (and also changes the high front vowel [i] to a back vowel [u] for an output [\textipa{"sulO}]. This output is likewise not attested in the training data. Finally, at 14900 steps, the network transforms the frication noise from a higher to lower kurtosis that corresponds to a labial fricative [f] in the training data ([\textipa{"fulO}]). At 20990 steps, it appears as if the network is introducing a period of aspiration noise and turning the fricative into a stop with the same following sequence [\textipa{"t\super hulO}]. None of these outputs are attested in the training  data, but the examples illustrate that the Generator searches for segment combinations with valid phonological processes in human language, such as \emph{devoicing}, \emph{occlusion}, or changing \emph{distribution of frication noise}.

\begin{figure}
\centering
\includegraphics[width=.8\textwidth]{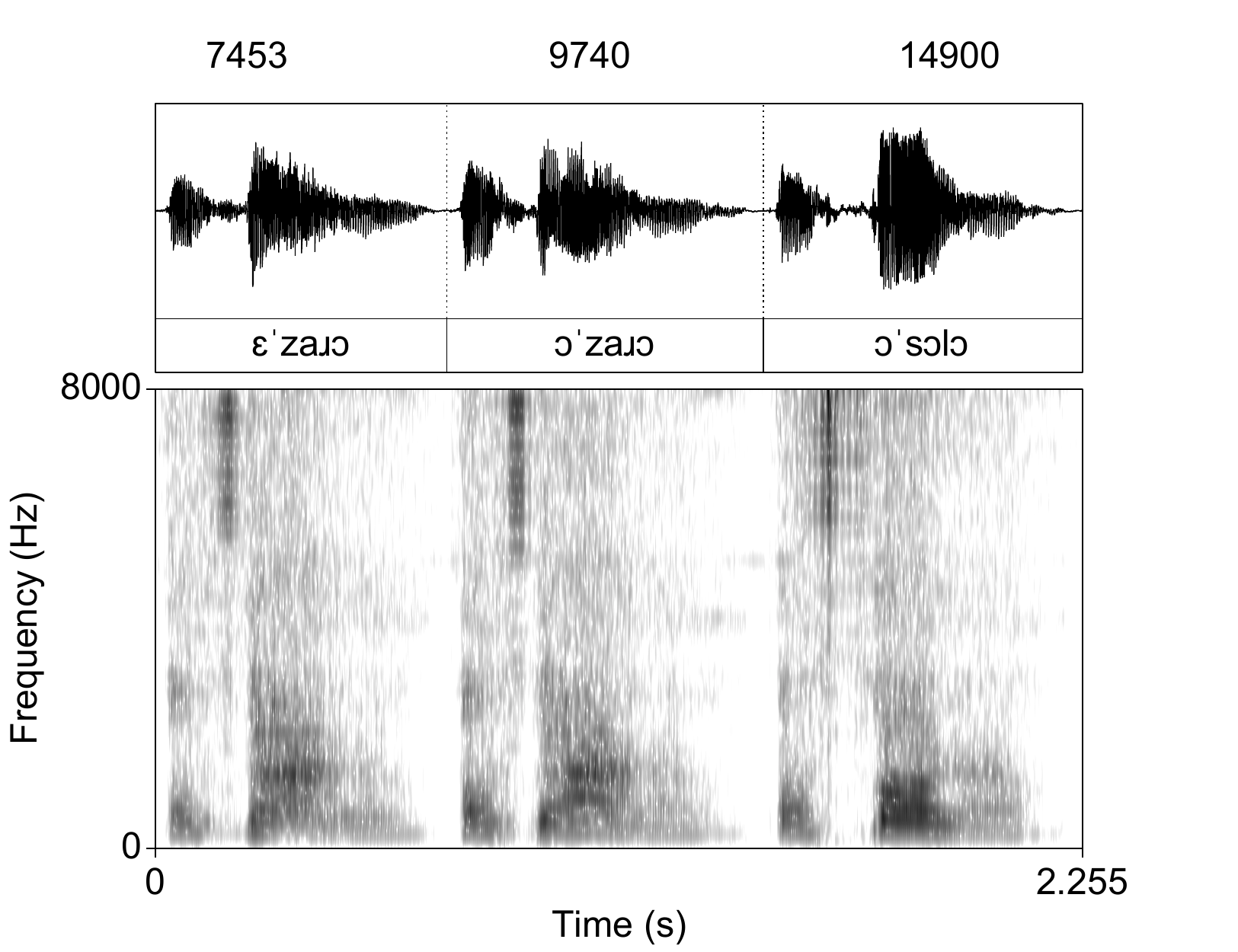}
\caption{\label{progressEzaroOsoroACL2020} Waveforms and spectrograms (0--8000 Hz) of three outputs illustrating changes in outputs with the same latent space values across different training steps (7453, 9740, and 14900).  } 
\end{figure}

Using this technique, we can not only observe how the network repairs distributional violations, but also how it searches through the space of possible segment combinations to repair violations of phonological rules in the data. Because the error rate of local phonological processes is relatively low in the output data, (1.8\% at 20990 steps), the study of how the network repairs outputs that violate phonological processes can only be exploratory at this point. An example that illustrates how learning progress can be directly observed with this method is given in Figure \ref{progressEzaroOsoroACL2020}. At 7453 training steps, the Generator outputs [\textipa{E"zA\*rO}] which violates both the local process of devoicing after a prefix and the non-local vowel harmony process. At 9740 steps, the second formant of the prefix vowel ([\textipa{E}]) substantially weakens and the formant structure of a back [\textipa{O}] emerges, which means the network repairs the harmony violation. At 14900 steps, voicing in the fricative ceases from the output, which means the output now conforms to the devoicing rule in the training data.  In other words, [z], which violates the phonological rule of devoicing after a prefix, devoices to [s], which conforms to the training data. At 14900 steps, the output thus fully conforms to the distributions in the training data: harmony and devoicing: [\textipa{O"sOlO}] (Figure \ref{progressEzaroOsoroACL2020}).  The output, while conforming to the rules of training data, is still innovative and none of the training inputs contains exactly this sequence. Spectrograms in Figure \ref{progressEzaroOsoroACL2020} illustrate how the network applies learning representations in its continuous outputs at different training steps that correspond to phonological processes in natural language: \emph{devoicing} and \emph{vowel-lowering}.

\subsection{Latent space}
\label{latentspace}

\begin{figure}
\centering
\includegraphics[width=.6\textwidth]{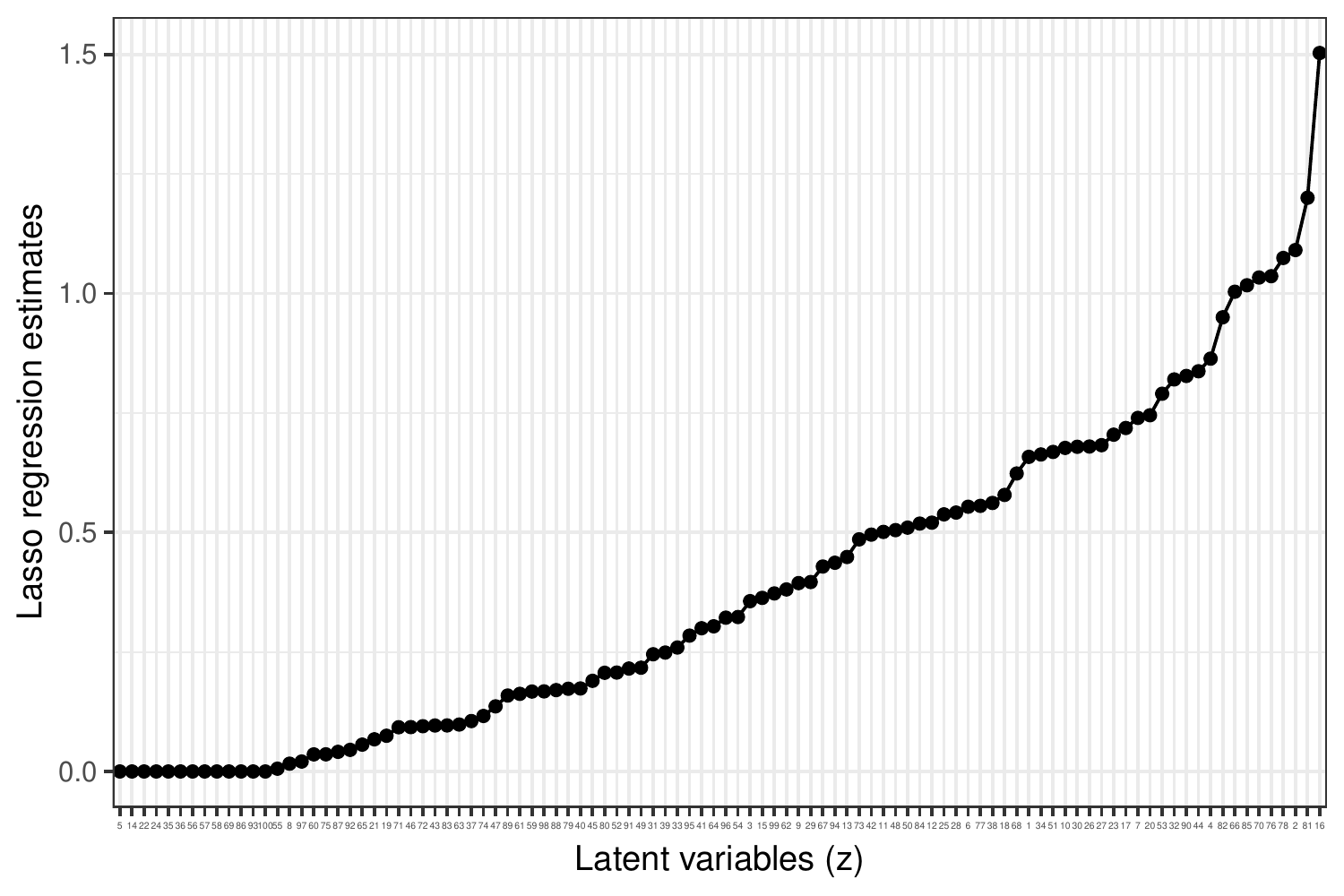}
\caption{\label{morphGANlassoPlot} Absolute Lasso logistic regression estimates  of a model with presence of the prefix as the dependent variable and values of 100 $z$-variables as predictors. The estimates are sorted in reversed order.} 
\end{figure}

To test how the network encodes prefixation in its latent space, we used a technique described in \citet{begus19} and Section \ref{materials} to identify dependencies between the latent space and generated data. 500 outputs of the Generator network trained after 20990 steps were transcribed and annotated for presence of the prefix V- and VN-.\footnote{All acoustic analyses are performed in Praat \citep{boersma15} by the author.} The number of steps for this analysis was chosen based on the analysis of progression of learning in Section \ref{progression}: it appears that a number of disharmonic outputs is repaired at  20990 steps and further training with more steps ceases to repair disharmonic outputs. That the network is successful in outputting data that approximates human speech in the training data is suggested by the fact that the author was unable to reliably transcribe the output in  only approximately 25 out of 500 outputs (5\%). The data were fit to a Lasso logistic regression model with the presence of the prefix as the dependent variable and the 100 latent variables of the Generator network as predictors (with the \emph{glmnet} package in \citealt{glmnet}). Alpha values were estimated with 10-fold cross-validation.  Estimates in Figure \ref{morphGANlassoPlot}  suggest that the network uses a single latent variable to encode the presence of the prefix in the output: there is a clear and substantial drop in regression estimates between $z_{16}$ and the rest of the latent space (other 99 $z$-variables). Such a substantial drop in regression estimates suggests that the network discretizes representation of the prefix into a single latent variable. 

To test the effect of $z_{16}$ on generated data, we generate 100 outputs with the value of $z_{16}$ set at $-4.5$ (for the method, see \citealt{begus19} and Section \ref{model}). Out of 100 generated samples, 100 (or 100\%) contain a prefix V- or VN-. When $z_{16}$ is set to its opposite value (4.5), only 1 out of 100 generated samples (1\%) contains a prefix.  This generative test suggests that the network encodes presence of the prefix in the output as a single variable in its latent space. By manipulating this feature, we can actively control the presence of the prefix in the output.\footnote{For a generative test showing that regression estimates indeed identify variables that correspond to a given phonetic/phonological representation, see \cite{begus19}.}

\subsection{Local and non-local processes}
\label{lnlp}

The training data contains evidence for local and non-local phenomena. Devoicing and occlusion after the prefixes V- and VN- are local; vowel harmony is non-local, as one or two segments intervene between the target and the corresponding vowel.

To test error rates of the output data, 500 outputs from the Generator networks trained after 20990 steps were analyzed. 211 outputs (42.2\%) were analyzed as involving a prefix VN- or V-. Of the 211 prefixed outputs, 162 (or 76.8\%) were analyzed as harmonious.\footnote{In one output excluded from the analysis, the prefix vowel is analyzed as [\textipa{A}].} Harmonious outcomes are consistently more frequent than non-harmonious both for front and back V$_2$ as well as across the two prefixes, V- and VN-. The distribution of the harmonious and disharmonious outputs across front and back triggering vowels and across the two prefixes are given in Table \ref{rawcounts}.

\begin{table}\centering
\begin{tabular}{lccccc}
\hline\hline
&\multicolumn{2}{c}{\textbf{VN-}}&\multicolumn{2}{c}{\textbf{V-}}&\textbf{Total}\\
&\textbf{front}&\textbf{back}&\textbf{front}&\textbf{back}&\\
\hline
\textbf{Harmonious}&53&31 &47&31&162\\
\textbf{Non-harmonious}&21&6&15&6&48\\
\textbf{\% Harmonious}&71.6\%&83.8\%&75.8\%&83.8\%&77.1\%\\

\hline\hline
\end{tabular}
\caption{\label{rawcounts}Raw counts of harmonious and disharmonious outputs of the Generator network across the two prefixes and vowel quality levels (front vs.~back). }

\end{table}

To test whether the Generator's higher rates of harmonious outcomes are significantly above chance, we fit the data to a linear logistic regression model with harmonious and non-harmonious outcomes as a dependent variable (harmonious coded as successes) and vowel \textsc{frontnesss} (with two sum-coded levels, front and back) and \textsc{prefix} identity (with two sum-coded levels, V- and VN-) as the independent variables with their interaction. Harmonious outcomes are significantly more frequent than disharmonious outcomes at means of all predictors: $\beta=  1.34,   z = 7.2, p<0.0001$. None of the interactions are significant. All estimates are given in Appendix Table \ref{genLinModel}. Predicted values of the model are plotted in Figure \ref{errorRateLogRegeffdfggplotBOTHPLOTS}. The results suggest that the network learns the non-local phonological process of vowel harmony, but imperfectly so: it violates the training data in approximately 23\% of outputs. The violations are  linguistically interpretable: the prefix vowel in the non-harmonious condition is not of random formant structure, but consists of formants characteristic of [\textipa{O}] or [\textipa{E}].

\begin{figure}
\centering
\includegraphics[width=.6\textwidth]{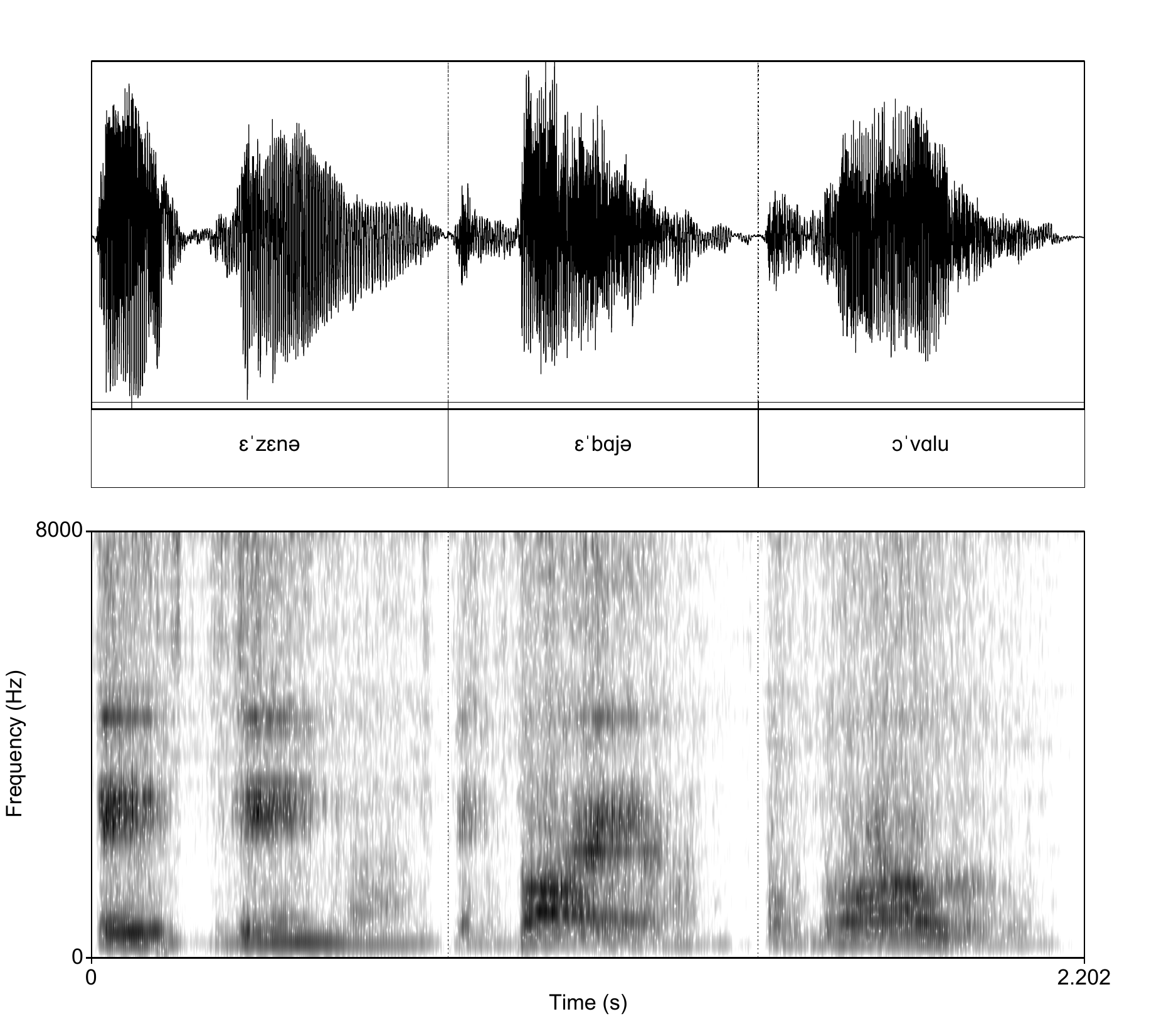}
\caption{\label{ezenaebajaovalu} Waveforms and spectrograms (0--8000 Hz) of three outputs of the Generator network trained after 20990 steps, [\textipa{E"\textbf{z}En@}], [\textipa{E"\textbf{b}Aj@}], and [\textipa{O"\textbf{v}Alu}], that violate the training data distributions with respect to local processes of fricative and stop devoicing.} 
\end{figure}

Local processes are substantially less frequent and easier to learn than non-local processes in natural languages. To test whether such distribution also emerges in deep convolutional networks, we can compare the error rate in the non-local process and the error rate in the local processes of the generated outputs. Out of 168 prefixed outputs containing a stop or a fricative, only three (1.8\%) violate the devoicing rule in the training data by which stops and fricatives are always voiceless in prefixed forms, e.g.~[\textipa{E"\textbf{z}En@}], [\textipa{E"\textbf{b}Aj@}],  and [\textipa{O"\textbf{v}Alu}] (spectrograms in Figure \ref{ezenaebajaovalu}).  This error rate is significantly lower compared to the error rate of the non-local process (OR =16.2 [5.1, 83.0], $p<0.0001$, Fisher Test). While the phonetic cues for harmony and devoicing are different and challenging to compare, it would be difficult to argue that the magnitude of phonetic cues for vowel formants (front vs.~back) is substantially smaller than the cue for voicing. The distribution aligns well with behavioral data in human subjects, where local processes have been shown to be  easier to learn than non-local processes in many studies \citep{finley11,finley12,mcmullin19,white18}.

\subsection{Emergence of rule-like behavior}
\label{erbb}

In the framework of symbolic representations,  vowel harmony can be derived with an algebraic rule (as in \ref{formalrule}). The harmony of the prefix vowel ([\textipa{E}]/[\textipa{O}]) is triggered by the following vowel V$_2$ via a rule that sets the feature [$\pm$front] in the  vowel of the prefix according to the value of the same feature in the following vowel (see formalism in \ref{formalrule}). Alternatively, the grammar can also operate on a morphophonological level: a prefix as a morphological unit can be chosen based on the value of the following vowel.

\begin{figure}
\centering
\includegraphics[width=.8\textwidth]{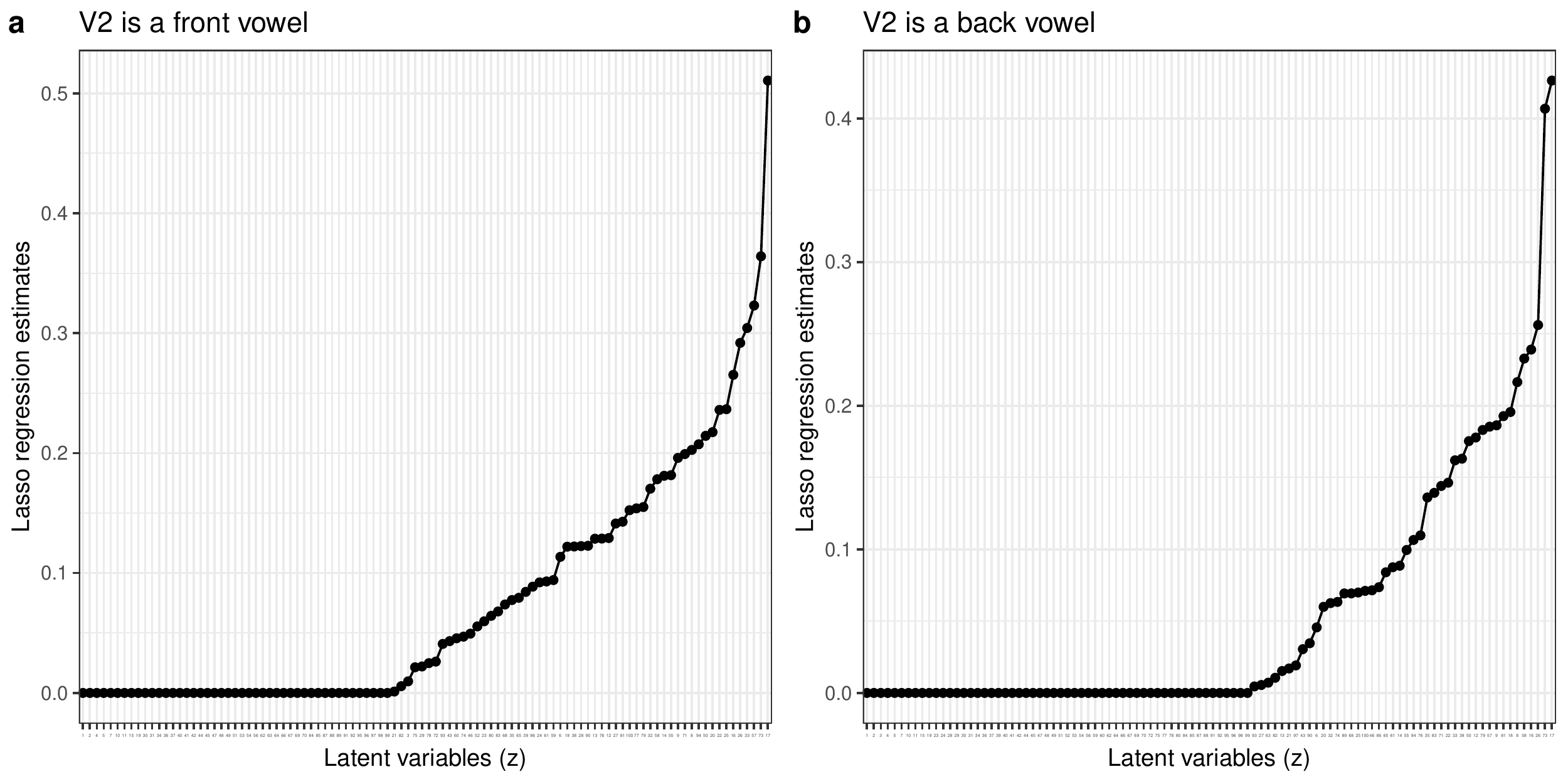}
\caption{\label{morphGANlassoPlotFrontBack} \textbf{(a)} Absolute Lasso logistic regression estimates  of a model with presence of front triggering vowels V$_2$ as the dependent variable and values of 100 $z$-variables as independent predictors. The estimates are sorted in reversed order. \textbf{(b)} Absolute Lasso logistic regression estimates  of a model with presence of back triggering vowels V$_2$ as the dependent variable and values of 100 $z$-variables as independent predictors. The estimates are sorted in reversed order.}
\end{figure}

We propose here that using the technique in \cite{begus19}, we can elicit such rule-like behavior in deep convolutional neural networks. The analysis in Section \ref{latentspace} suggests that the Generator learns to associate $z_{16}$ with presence of a prefix. There is a substantial drop in regression estimates after the estimates for $z_{16}$, which suggests that the network discretizes the continuous phonetic input  and uses a single variable to encode presence of some phonetic/phonological material which corresponds to a morphological unit: a prefix. To elicit rule-like behavior, we can identify another variable in the latent space --- the variable that corresponds to the frontness/backness of vowel V$_2$. To identify such a variable, the generated 500 outputs are annotated for vowel (V$_2$) frontness. We fit the data to two linear logistic regression models: one in which outputs with the front vowel (V$_2$) [\textipa{E, i}] are coded as success and another in which [\textipa{A, O, u}] are coded as success. The independent variables are values of the 100 latent variables $z$ randomly sampled for each of the 500 annotated generated outputs. The model is fit using the \emph{glmnet} package \citep{glmnet} in R \citep{r}.  Lambda values are estimated with 10-fold cross-validation. Estimates of the two models are given in Figure \ref{morphGANlassoPlotFrontBack}.

Both models uniformly suggest that $z_{17}$ is the latent variable most strongly associated with determining vowel frontness of the triggering vowel V$_2$. Regression estimates again suggest that the Generator network learns to encode vowel frontness with a single latent variable: there is a substantial drop of estimates after the single latent variable $z_{17}$. Negative values of $z_{17}$ correspond to presence of front [\textipa{E, i}] in V$_2$, while positive values correspond to presence of back [\textipa{A, O, u}] (estimates in Figure \ref{morphGANlassoPlotFrontBack} are in absolute values). 

To elicit rule-like behavior, we force the prefix in the input and simultaneously force vowel V$_2$ to turn from a front vowel [\textipa{E, i}] into a back vowel [\textipa{A, O, u}]. To achieve this affect, we simultaneously  manipulate $z_{16}$ (presence of prefix) and  $z_{17}$ (frontness of vowel).\footnote{That the two variables are consecutive is likely a coincidence.} If the Generator network learned vowel harmony, then the vowel of the prefix should change together with the forced change of vowel quality. Such a behavior would parallel rule-based computation: setting a single variable to a value that forces prefixation in the output and manipulating the variable that changes the conditioning environment (V$_2$) results in a process that changes the target vowel according to the condition --- vowel harmony.

To test this hypothesis, we set the value of $z_{16}$ to $-2.5$ which forces the prefix in the output. Additionally, we generate outputs with $z_{17}$ interpolated from values $-6$ to 6 in increments of 1. 60 such sets of 13 generated samples (with $z_{17}$ from $-6$ to 6) are generated and acoustically analyzed (780 outputs total). That $z_{16}$ indeed causes the prefix in the output is suggested by the count of prefixed forms in the output: 635 out of 780 generated samples  (or 81.4\%) were analyzed as featuring a prefix (for an independent test of the effect of $z_{16}$ on presence of prefix, see Section \ref{latentspace}).

\begin{figure}
\centering
\includegraphics[width=.9\textwidth]{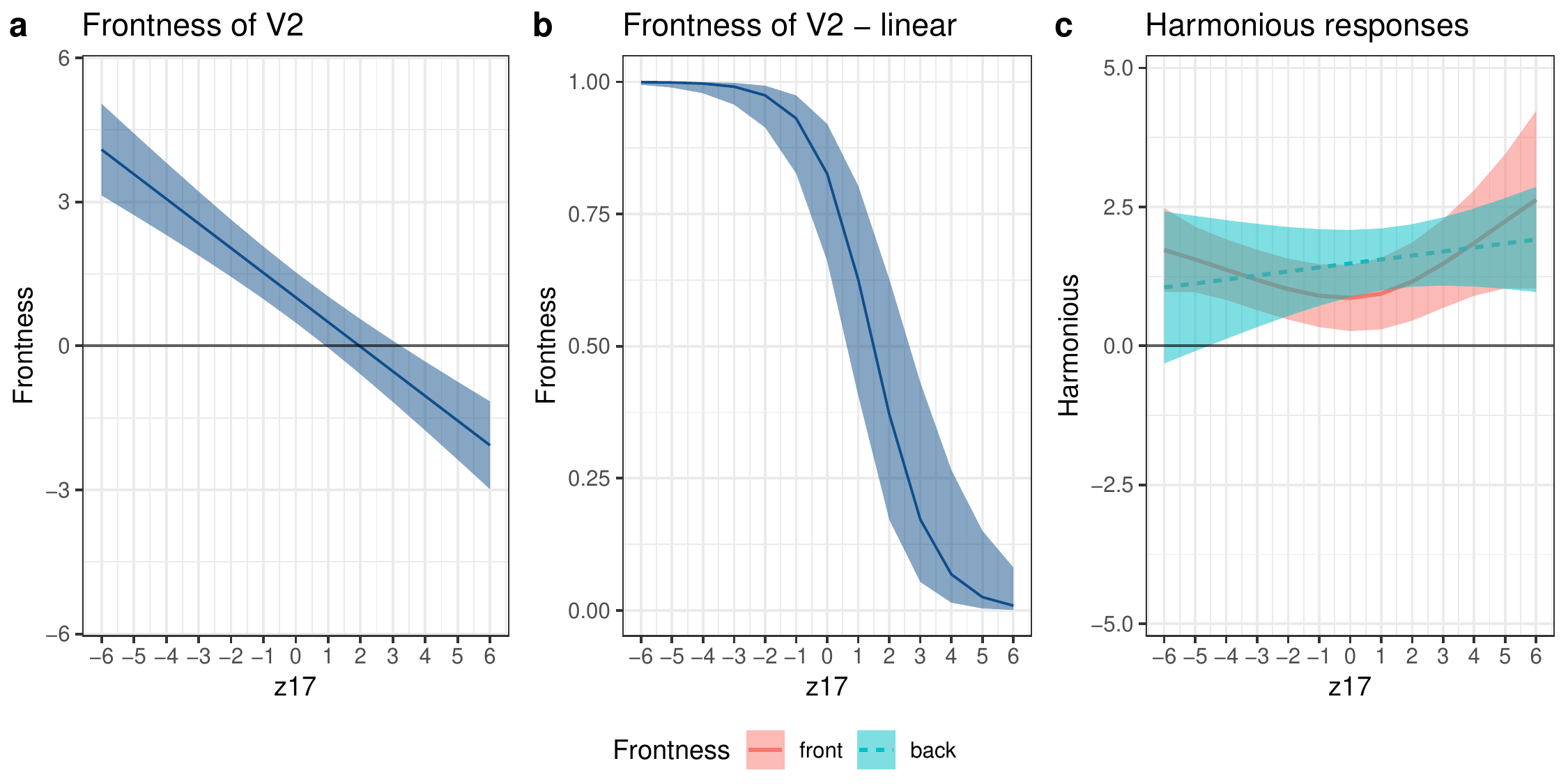}
\caption{\label{zLassoMGANz16z17bamBOTHPLOTS} \textbf{(a)} Fitted values and 95\% CIs of a generalized additive mixed effects logistic regression model with the front vs.~back triggering vowel (V$_2$) value as the dependent variable and thin-plate smooths for values of $z_{17}$ as the independent variable (with random smooths for each of the 60 generated sets). The estimates show that $z_{17}$ causes a change from a front to a back vowel as its values are interpolated from $-6$ to 6 and that the relationship between values of $z_{17}$ and frontness/backness of the vowel are linear. The regression estimates are in Appendix Table \ref{interpolation1}.  \textbf{(b)} Fitted values and 95\% CIs  of a linear mixed effects logistic regression model with the front vs.~back triggering vowel (V$_2$) value and random intercepts and slopes for each of the 60 trajectories. The plot illustrates how the percent of front vowels decreases as the value of $z_{17}$ increases (and vice-versa for back vowels). \textbf{(c)} Fitted values and 95\% CIs  of a generalized additive mixed effects logistic regression model with harmonious (success) and disharmonious (failure) outcome as the dependent variable, vowel \textsc{frontness} as a parametric predictor, and thin-plate smooths for the two levels of frontness (front vs.~back) across the values of $z_{17}$ and random smooths for each of the 60 set of generated outputs. Estimates of the model are given in Appendix Table \ref{interpolation2}.} 
\end{figure}

That $z_{17}$ indeed changes the triggering vowel V$_2$ from a front [\textipa{E, i}] to a back  [\textipa{A, O, u}] is strongly suggested by the generated outputs. We annotate the 635 prefixed forms from the 60 sets of generated interpolated outputs for frontness and backness of the triggering vowel V$_2$. We fit the annotated data to a generalized additive mixed logistic regression model (GAMMs; \citealt{mgcv}) with an intercept and thin-plate smooths that estimate how the presence of a front or back vowel in the output changes with interpolated values. A random smooth for each trajectory (each of the 60 generated sets) is added to the model (estimates in Table \ref{interpolation1}). Figure \ref{zLassoMGANz16z17bamBOTHPLOTS} suggests that the presence of $z_{17}$ causes the triggering vowel from a front one at values in the negative range to a back one at positive values. The relationship appears to be linear even when the model does not have an assumption of linearity (GAMM).  If we refit the data to a linear logistic mixed effect regression (with a random intercept for trajectory and by-trajectory random slopes), we get a significant negative correlation between values of $z_{17}$ (from $-6$ to 6) and percent of front vs.~back output ($\beta =   -1.04, z=-5.38,p< 0.0001$). Figure \ref{zLassoMGANz16z17bamBOTHPLOTS} illustrates how rates of front vowel V$_2$ in the output change from almost 100\% at one end of spectrum to 0\% (or 100\% of back vowel) in the other end of spectrum.

\begin{figure}
\centering
\includegraphics[height=.25\textheight]{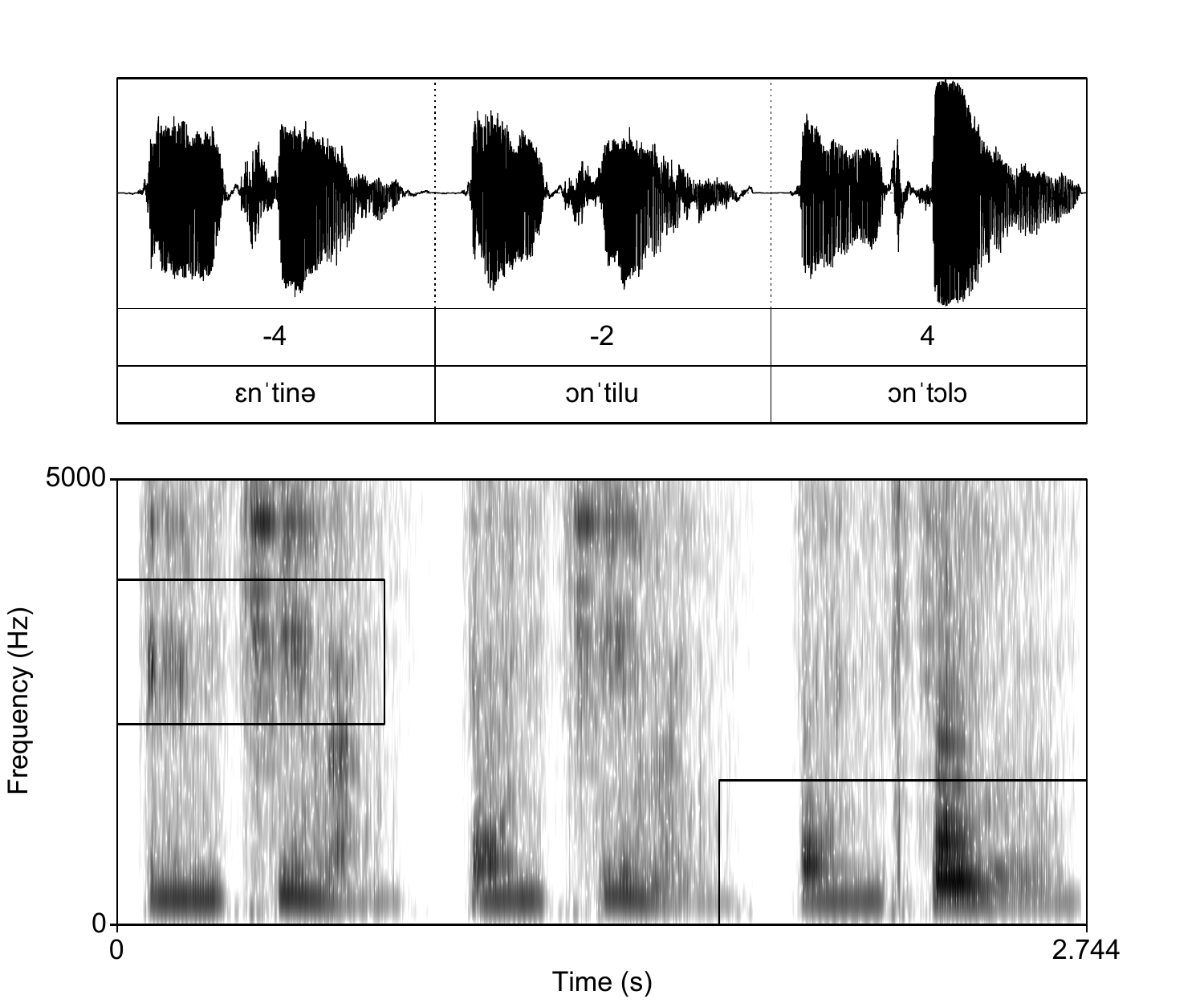}\includegraphics[height=.25\textheight]{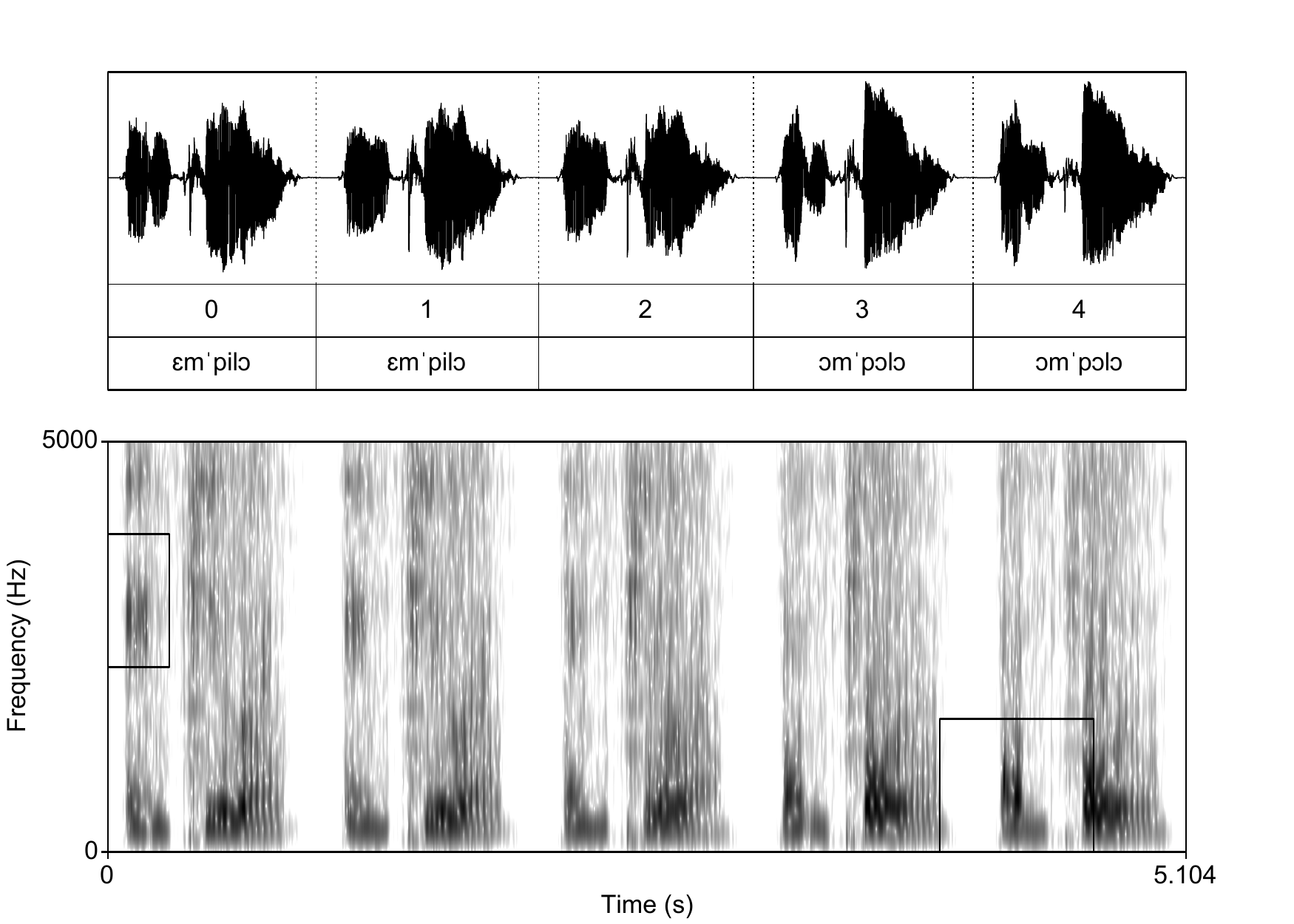}
\caption{\label{z57} Waveforms and spectrograms (0--5000 Hz) of outputs with  interpolated values of $z_{17}$ that change the triggering vowel V$_2$ from a front [\textipa{E, i}] to a back [\textipa{A, O, u}] and $z_{16}$ set at $-2.5$, which forces the prefix in the output. \textbf{(left)} Three outputs with $z_{17}$ set at $-4, -2,$ and $2$. The spectrogram shows how the formant structure of a front [\textipa{E}] in the prefix changes to the formant structure of a back [\textipa{O}] as the triggering vowel changes from a front [i] to a back [\textipa{O}]. \textbf{(right)}  Five outputs with $z_{17}$ set at $0, 1, 2, 3,$ and $4$. The spectrogram again shows an automatic change of the prefix vowel consistent with the vowel harmony in the training data. Areas in squares indicate formant structures of interest. } 
\end{figure}

 To test whether the prefix vowel is harmonious even when the variable changing the triggering vowel is interpolated, we annotate the  635 prefixed forms from the 60 sets for frontness of the triggering vowel V$_2$ and for vowel harmony. Data is annotated for harmony (successes vs.~failures) and fit to a generalized additive mixed effects logistic regression model. The independent variables are \textsc{frontness} of the vowel (treatment-coded with back as reference) and a thin plate smooth for values of $z_{16}$ as well as by-trajectory random smooths (estimates in Table \ref{interpolation2}). The estimates of the parametric term suggest that the prefix vowel is harmonious both for front and back triggering vowels V$_2$. Harmonious outputs with a back triggering vowel V$_2$ ([\textipa{A, O, u}]) are significantly more frequent that non-harmonious outputs: $\beta =1.43, z=   4.23, p< 0.0001$. That the same is true for the front vowel is clear from estimates in Figure \ref{zLassoMGANz16z17bamBOTHPLOTS} (confidence intervals do not cross zero) and from the fact that estimates for the front triggering vowel V$_2$ are not different from estimates for back vowel. This is confirmed if we refit the model with sum-coded \textsc{frontness} factor ($\beta=1.41,z=   6.30, p<0.0001$). We also observe a slight negative trend in harmonious outcomes as we increase $z_{17}$ and a slight positive trend for harmony in the back vowel conditions, although estimates for smooths are not significant. This likely results from the trend that we observe in the data: as we force the triggering vowel to be front  (by setting z$_{17}$ to $-6$), the prefix is harmonious. When the vowel changes as we interpolate the value of $z_{17}$, we have a higher proportion of disharmonious outputs, because apparently the underlying value of the triggering vowel is not ``strongly'' front or back. As the value of $z_{17}$ increases towards 6 and the back vowel is forced more strongly in the output, we get a higher proportion of harmonious outputs again (of course with a back vowel harmony).\footnote{While the estimates of the effects are significant, the trends are not categorical. Occasionally, the vowel does not change from front to back (or from non-harmonious to harmonious) and more rarely, trends are reversed.}  Figure \ref{z57} illustrates the gradual change of the forced prefix from a front (containing an [\textipa{E}]) to back (containing an [\textipa{O}]) when $z_{17}$ changes the vowel V$_2$ from a front to a back vowel.  In other words, as we force a change of the triggering vowel quality from front to back with a single latent variable, the prefix (also forced with a single variable) automatically changes in order to remain harmonious.
 
The deep convolutional network thus appears to represent what would approximate a rule-like computation in phonology: as we force the prefix in the output and change the quality of the triggering vowel from front to back by manipulating only two latent variables, vowel harmony emerges automatically. The appearance of rule-based computation is not categorical --- but as is always the case in connectionism, probabilistic  --- as the prefix does not always change to be harmonious and other features can change along the observed changes. This is to the author's knowledge the closest approximation of rule-based phenomena, especially considering that the models contain no language-specific mechanism and are trained in an unsupervised manner from raw acoustic data.

It is possible that the emergence of rule-like behavior results from the choice of distribution of $z$-variables or other hyperparameters in the model.  For example, $z$-variables can take a variety of distributions, from uniform, Gaussian, to Bernoulli distributions. Testing how hyperparameters influence behavior of the models and what implications this can bring for cognitive modeling are left for future work. A related experiment, however, in which the latent variables have Bernoulli distributions show a very similar behavior when tested on another morphophonological process --- reduplication \citep{begusCiw}. In the present experiment, $z$-variables are uniformly distributed in the interval $(-1,1)$. In an experiment testing reduplication \citep{begusCiw}, a subset of latent variables (code variables) are Bernoulli distributed (0 or 1) that constitute a one-hot vector. Even with this distribution, interpolation and setting variables to marginal values outside of the training interval result in a rule-like behavior and a near one-to-one correspondence between the Bernoulli distributed variables and an identity-based morphophonological pattern.\footnote{The experiment in \cite{begusCiw} is trained on an InfoGAN extension \citep{chen16} where another network is introduced that forces the Generator to output informative data.} Future work should test the effects of normally distributed variables and other hyperparameters, such as the number of convolutional layers and the number of latent variables.

\section{Paralleling neural networks and artificial grammar learning experiments}
\label{paralleling}

\begin{figure}
\centering
\includegraphics[width=.5\textwidth]{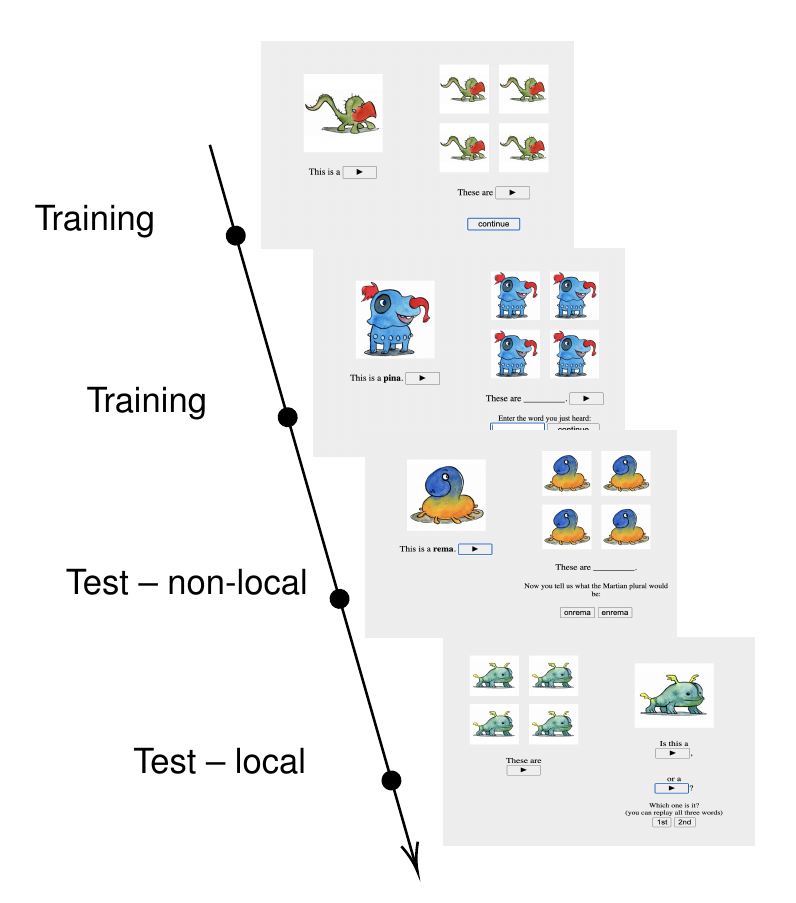}

\caption{\label{expdesMGAN} Experimenal design (from \citealt{begusCatalysis}) in the Experigen interface \citep{becker13}; artwork from \cite{vijver14}. The order of the training and the test phases are randomized, but the training precedes the test block. For the exact procedure of the experiment, see \citealt{begusCatalysis}.} 
\end{figure}

To parallel the performance of the computational experiment with results from a behavioral experiment, we combine novel data presented here for the first time with results of an experiment in \cite{begusCatalysis}.  The subjects were trained on the same data as used in the computational experiment, but divided into two separate experiments: one in which subjects were trained on data with the VN- prefix and another one on data with the V- prefix. Subjects were recruited via Amazon MTurk\footnote{That the results of the experiment are not heavily influenced by the participants in the behavioral experiments being recruited via Amazon MTurk is suggested by the fact that vowel harmony outcomes are very similar to a related experiment with similar training data that was performed in-person with the supervision of a research assistant in which subjects were recruited from the general public \citep{begusCatalysis}.}, completed informed consent before participating, and were presented with experimental stimuli in Experigen \citep{becker13}. In the behavioral experiment, the unprefixed-prefixed forms are presented to subjects in pairs, where the prefixed form carries the function of plural. Subjects were presented with a picture of a Martian creature. A single creature is associated with the unprefixed form; four creatures are associated with the prefixed form. The experimental interface is illustrated in Figure \ref{expdesMGAN}.

Subjects whose first language was not English or who had self-reported linguistic education were removed from the analysis. Altogether 333 subjects that provided 1987 responses on the vowel harmony test are analyzed \footnote{For detailed discussion on exclusion criteria, see \cite{begusCatalysis}. In the V- condition, we excluded participants with non-unique  Amazon MTurk IDs as well as with those IDs who had already taken the VN-experiment.}

The training phase in the VN- experiment consisted of 58 pairs of bare and prefixed forms. All examples were harmonious and some included evidence for the local processes of post-nasal devoicing and post-nasal devoicing and occlusion (as described in detail in the Section \ref{data} on data used in the computational experiment). In the V- experiment, the training phase consisted of 60 pairs of bare and prefixed forms, all of which contained evidence for harmony and some of which contain evidence for local processes of devoicing and devoicing and fricativization  (see Section \ref{data}). All items used in the behavioral experiment are listed in Appendix Tables \ref{nonce1}, \ref{nonce1a}, \ref{nonce2}, \ref{nonce2a}, \ref{nonce3}, \ref{nonce3a}, and \ref{test3a}. 

After the training phase, the subjects were tested on six bare forms with C$_1$ either a [r] or [l] (three with a front V$_2$ and three with back) and had to choose between harmonious and non-harmonious responses in a forced choice task (see Test -- Local in Figure \ref{expdesMGAN}), as well as between various local processes. For example, subjects were presented with a stimulus [\textipa{"lirO}], presented auditorily and orthographically, and had to choose between the plural form \emph{eliro} (harmonious) and  \emph{oliro}, presented only orthographically.\footnote{In the test phase on local processes involving the prefix VN-, the subjects were presented with a plural form exclusively auditorily and had to choose between two possible singular forms: one consistent with devoicing and another consistent with devoicing and occlusion. In the V- condition, the subjects similarly chose between singular forms consistent with intervocalic devoicing or intervocalic devoicing with fricativization.}

While the behavioral experiments do not directly test whether non-local processes are more difficult to learn than local processes (this has already been confirmed experimentally in several studies; see \citealt{finley11,finley12,mcmullin19,white18}), the local process is made more difficult to learn in the experiment: subjects were explicitly instructed to learn the (non-local) distribution of prefixes (vowel harmony), but never about learning the local processes. Moreover, the learning of local processes is tested exclusively with auditory stimuli. 

To test the learning of the non-local process in the behavioral experiment, the responses were fit to a linear mixed effects logistic regression model (\emph{lme4} package by \citealt{lme4}). First, we fit the full model with harmonic vs.~non-harmonic responses (successes vs.~failures) as the dependent variable and \textsc{frontness} (front vs.~back, sum-coded) of the vowel and the shape of the \textsc{prefix} (VN- vs.~V-, sum-coded) as the independent variable (with interaction) and random intercepts for \textsc{subject} and \textsc{item} with by-subject and by-item random slope for \textsc{frontness}. The final model was chosen based on Akaike Information Criterion (AIC) by removing random slopes first and then interactions. The final model includes the \textsc{frontness $\times$ prefix} interaction and random intercepts for \textsc{subject} and \textsc{item}. 

The results show that subject learn the vowel harmony pattern from the training data ($\beta = 0.56,z=  5.0, p < 0.0001$).  In other words, harmonious responses are significantly above the chance level, which suggests subjects do learn the harmonious pattern. However, the error rate is quite high. The 95\% profile CIs for the preference for harmonious response are quite low: [57.6\%, 69.2\%], especially given that 234/270 items are bare-prefixed pairs each of which contains evidence for vowel harmony. All regression estimates are in Table \ref{subjectsExperiment}. 

\begin{figure}
\centering
\includegraphics[width=.6\textwidth]{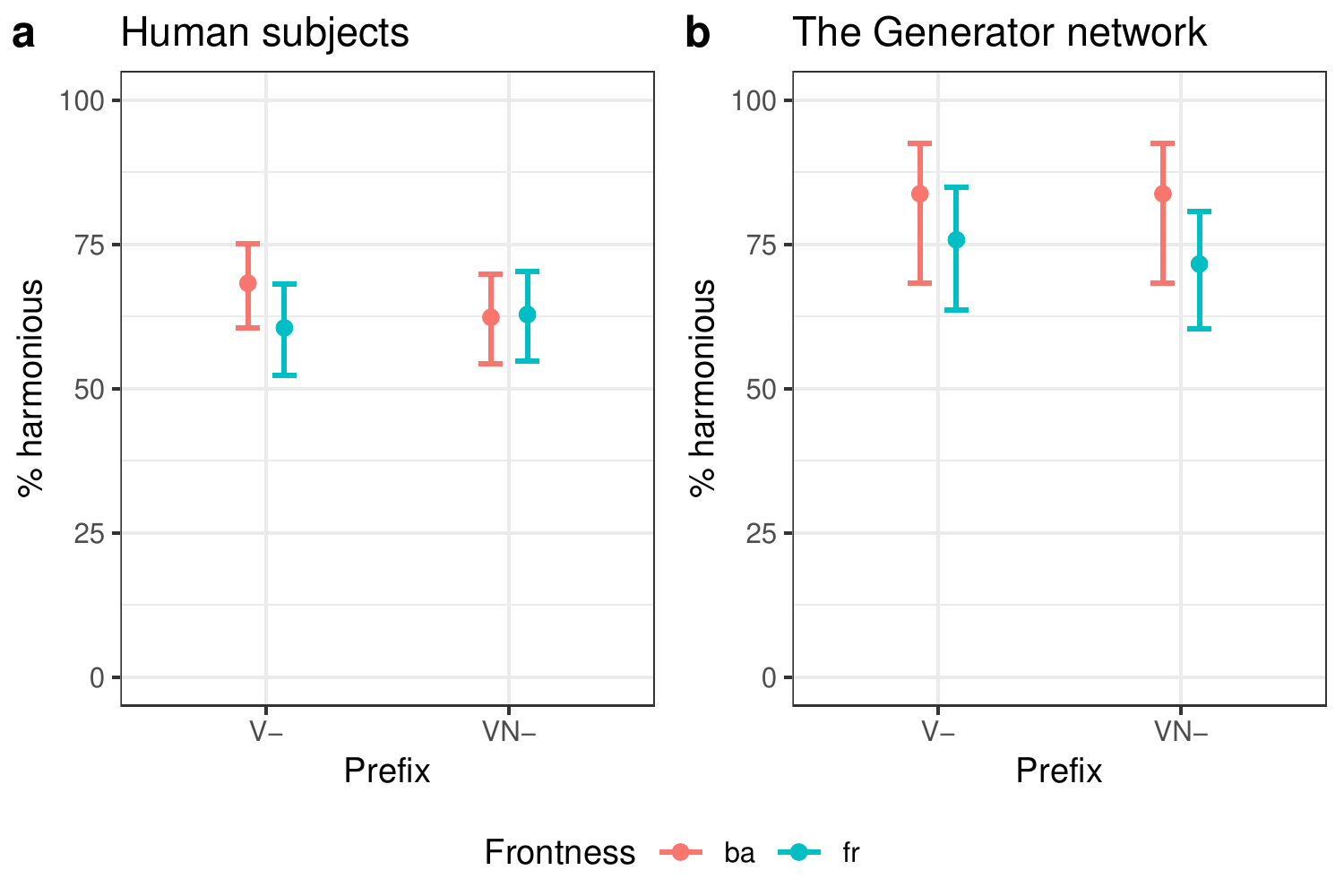}

\caption{\label{errorRateLogRegeffdfggplotBOTHPLOTS} \textbf{(a)}  Estimates of the linear mixed effects logistic regression model with harmonious responses of human subjects in the behavioral experiment as successes and vowel frontnesss and prefix identity as the independent variables with their interaction.  \textbf{(b)} Estimates and 95\% CIs of the linear logistic regression model with harmonious outcomes of the Generator network as successes and vowel frontness and prefix identity as the independent variables with their interaction.}
\end{figure}

We can directly compare subject's responses in the behavioral experiments with outputs of the computational experiment. The Generator network violates local distributions in the data in only three out of 168 generated outputs with a prefix and a stop or a fricative (1.8\%). On the non-local task, however, the Generator's error rate is substantially higher and similar to the error rate in the artificial grammar learning experiment conducted on human subjects. Figure \ref{errorRateLogRegeffdfggplotBOTHPLOTS} illustrates the similarity.

To be sure, there are substantial differences between the computational and behavioral experiment. First, the comparison is necessarily superficial, because this paper does not claim that humans learn phonological patterns in the same way as deep convolutional networks; however, this does not preclude us from comparing their performance. The number of epochs in the computational experiment is $\sim$ 24877, while subjects were only exposed to training data once. On the other hand, human subjects were adults with full language capacity and already established phonological inventories, phonological grammar, and articulatory and perceptual mechanisms. The Generator network has to learn to produce speech-like outputs from random noise and does not contain any language-specific learning mechanisms. 

This comparison in performance between human subjects and the computational model suggests that non-local processes are computationally similarly costly both for humans and for computational models of language acquisition to the degree that the error rates across the two conditions are similar. That non-local processes are computationally costly has of course been shown before, but to our knowledge, this is the first such confirmation on a deep convolutional neural network model that is trained on the same data as human subjects and that learns speech representations from raw acoustic data.

\section{Discussion}

This paper tests learning of local and non-local processes in human speech with deep convolutional networks in the GAN architecture. More specifically, we test the learning of non-local vowel harmony and local devoicing processes in a setting that approximates morphological and phonological processes in language: the model is trained on raw speech data with bare and prefixed forms in random order.

First, we argue that deep convolutional GANs output highly informative data despite being trained on extremely small datasets (N = 270) with a high number of epochs. The outputs are acoustically analyzable and linguistically interpretable. The Generator learns local processes and phonotactic restrictions with low error rates which suggests that training is successful for at least a subset of training objectives. As has been shown before \citep{begus19,begusCiw}, however, the Generator also outputs innovative data that violate training data. These violations are not random, but are linguistically interpretable.  23.2\% of outputs are disharmonious, and 33.7\% are innovative outputs (harmonious or unprefixed)  that conform to phonotactic and distributional properties of the training data, but include unique sequences that are never present in the training data (Section \ref{small}). In only $\sim$5\% of annotated outputs is the data not linguistically interpretable. Innovative outputs also suggest that the Generator does not overfit despite the high number of epochs, in line with previous work on overfitting in GANs. The finding that GANs can be trained on very small data sets should open up several new possibilities for research on deep convolutional networks, speech, and internal representations in deep convolutional networks.

An exploratory study of innovative outputs suggests that, in order  to repair its data violations, the network uses strategies that approximate processes in human phonology: devoicing, occlusion, and distribution of frication noise. We propose that these repairs can be directly followed with progression of learning by keeping the random latent variables constant while generating data from the network trained at different training steps. Acoustic analysis of outputs at different training steps in Section \ref{progression} identifies strategies that the network uses to repair violations in data distributions.

One of the objectives of this paper is to explore how deep convolutional networks trained in the GAN framework on raw speech discretize linguistically meaningful representations in the latent space, especially with respect to non-local morphophonological processes. The raw acoustic data hearing human infants are faced with is continuous. Phonological computation discretizes the continuous space into discrete representations and manipulates these representations, which results in phonological processes such as vowel harmony. Using the technique in \cite{begus19}, we identify variables in the Generator's latent space that correspond to linguistically meaningful units, such as presence of a prefix or frontness of a vowel. Lasso regression estimates suggest that the network uses a minimal number of variables to represent presence of a prefix in the output. In other words, the steep drop in the regression estimates after the variable with the highest estimate suggests that the network discretizes some continuous phonetic content in its internal space.  The same is true for a phonetic feature such as frontness of the first vowel in bare forms (V$_2$). The network appears to primarily use a single variable to encode this phonetic property of outputs. An independent generative test suggest that manipulating this one variable  on a linear scale well outside the training range (from $-6$ to 6) results in a gradual and linear transition from a front to a back first vowel (Figure \ref{zLassoMGANz16z17bamBOTHPLOTS}).

This paper argues that an approximation of a symbolic rule emerges as an interaction between latent variables in deep convolutional networks. To test learning of the non-local vowel harmony, we force a prefix in the output with a single variable ($z_{16}$ at $-2.5$) and force the change of the triggering vowel from front to back with a linear interpolation of a single variable ($z_{17}$).  The statistical tests in Section \ref{erbb} suggest that the generated outputs remain harmonious in the majority of cases despite the change of the triggering vowel. In other words, the rule-like vowel harmony emerges automatically in a deep convolutional network from an interaction of the variable that forces some morphophonological entity in the output (the prefix) and the variable that changes the triggering segment. While harmonic outputs are significantly more frequent than non-harmonic outputs, the distribution is probabilistic rather than categorical. Another trend emerges from the statistical tests: the outputs are more likely to be non-harmonic in the transition period when the triggering vowel changes from front to back. It is likely the case that the relative strength of frontness and backness affects the rates of harmonic vs.~non-harmonic outcomes. In other words, it appears that the prefix harmony is not triggered until the frontness/backness feature of the triggering vowel is strong enough, i.e.~has a high enough latent variable value. That phonological features bear inherit weights (that can be conceptualized as strength or latent variable values in our model) has been argued before in the Optimality Theoretic framework \citep{smolensky16,smolensky19}.

Phonological computation has been shown to favor local processes over non-local processes. Many studies show experimentally that the learning of non-local processes is more difficult \citep{finley11,finley12,mcmullin19,white18}. This learning bias is also reflected in  typology: the majority of phonological processes are local in the world's languages \citep{finley11}. A clear preference for locality emerges in our computational experiment as well: despite substantially more evidence for the non-local process in the training data, the error rate is significantly higher in the non-local condition in the Generator's network. Whether the prevalence of some patterns in human speech results from articulatory factors (e.g.~the articulation of sounds is most strongly affected by the immediately preceding or following sounds) or from learnability (e.g.~the learning of non-local processes is more difficult) has been a focal topic of discussion in phonology, linguistics, and cognitive science in general.  While this result does not offer an answer as to whether the preference for non-locality in typology results from learning or a language's cultural transmission \cite{begusCatalysis}, it does provide evidence that non-locality preferences can be explained with domain-general cognitive mechanisms using deep neural networks.

It is possible that the Generator network violates the non-local vowel harmony relatively frequently (in 23.2\% of the outputs) because it is not fully trained and potentially converges on a local optimum. Even if this is the case, the results are nevertheless informative for our objectives. First, the Generator is clearly well trained on the local processes: error rate for the local process of devoicing is 1.8\%. Second, the Generator is well trained on the phonotactic restrictions in the training data: the error rate for the phonotactic restrictions is 0\% if we exclude unanalyzable outputs (constituting only 6.5\% of the outputs). Since our primary objective is to compare the learning of local and non-local processes in speech, the fact that local processes are well learned, and significantly better compared to the non-local process (see Section \ref{lnlp}), suggests that non-local processes are more difficult to learn than local processes in deep convolutional networks in the GAN framework. Finally, this paper illustrates the importance of analyzing the models at different training steps (as proposed in Section \ref{progression}) when the primary objective is probing learning representations, neural network interpretability, cognitive modeling, or linguistic relevance of the models. One of the potential concerns in fully trained models is the so-called ceiling effect. If the model were able to perform equally well on both local and non-local processes, we might erroneously conclude that local and non-local processes are equally learnable, whereas one could have been learned substantially earlier in the training than the other.

Because GANs trained on small datasets produce informative results, we can use the same stimuli for training deep convolutional networks and artificial grammar learning experiments on human subjects. We compare data from a behavioral experiment that tested the learning of vowel harmony. Results show a similar degree or error rate across the computational and artificial grammar learning experiments. It is true that the Generator network does not output vowel harmony categorically (as opposed to local processes, which are near categorical), but neither do the human subjects tested in a behavioral experiment perform at the categorical level. 
This suggests that non-local processes are, from a learnability viewpoint, similarly costly both for the deep convolutional network and for human subjects.

\section{Conclusion}

The results of the present experiment provide new information on internal representations in deep convolutional networks trained on raw speech, and bear evidence for the long-standing discussion on symbolism vs.~connectionism in cognitive science. The networks not only represent morphophonological units with discretized representations (resembling the morphological level), but also learn to encode morphophonological processes (resembling rule-like computation). An approximation of rule-like non-local generalizations in the data emerges from training a deep convolutional GAN. We provide evidence arguing that human behavioral data superficially matches the outcomes of the computational model. Applying such an experiment to further data should yield a clearer picture on how rule-like generalizations emerge as interactions between variables in deep convolutional neural networks trained on raw speech data, and how performance and biases of deep neural networks corresponds to human performance in behavioral experiments.

\subsubsection*{Acknowledgements}
This research was funded by a grant to new faculty at the University of Washington and UC Berkeley as well as by Harvard Mind Brain Behavior and Department of Linguistics.

\bibliographystyle{elsarticle-harv}

\appendix

\section{Appendix}

\subsection{Training data}
The recordings or training data were made in a sound-attenuated booth at the Department of Linguistics at Harvard University using a USBPre 2 (Sound Devices) pre-amp and Shure 53 Beta omnidirectional condenser head-mounted microphone in Audacity (originally sampled at 44.1 kHz and then downsampled to 16 kHz). 

\begin{figure}[H]
\centering
\includegraphics[width=.5\textwidth]{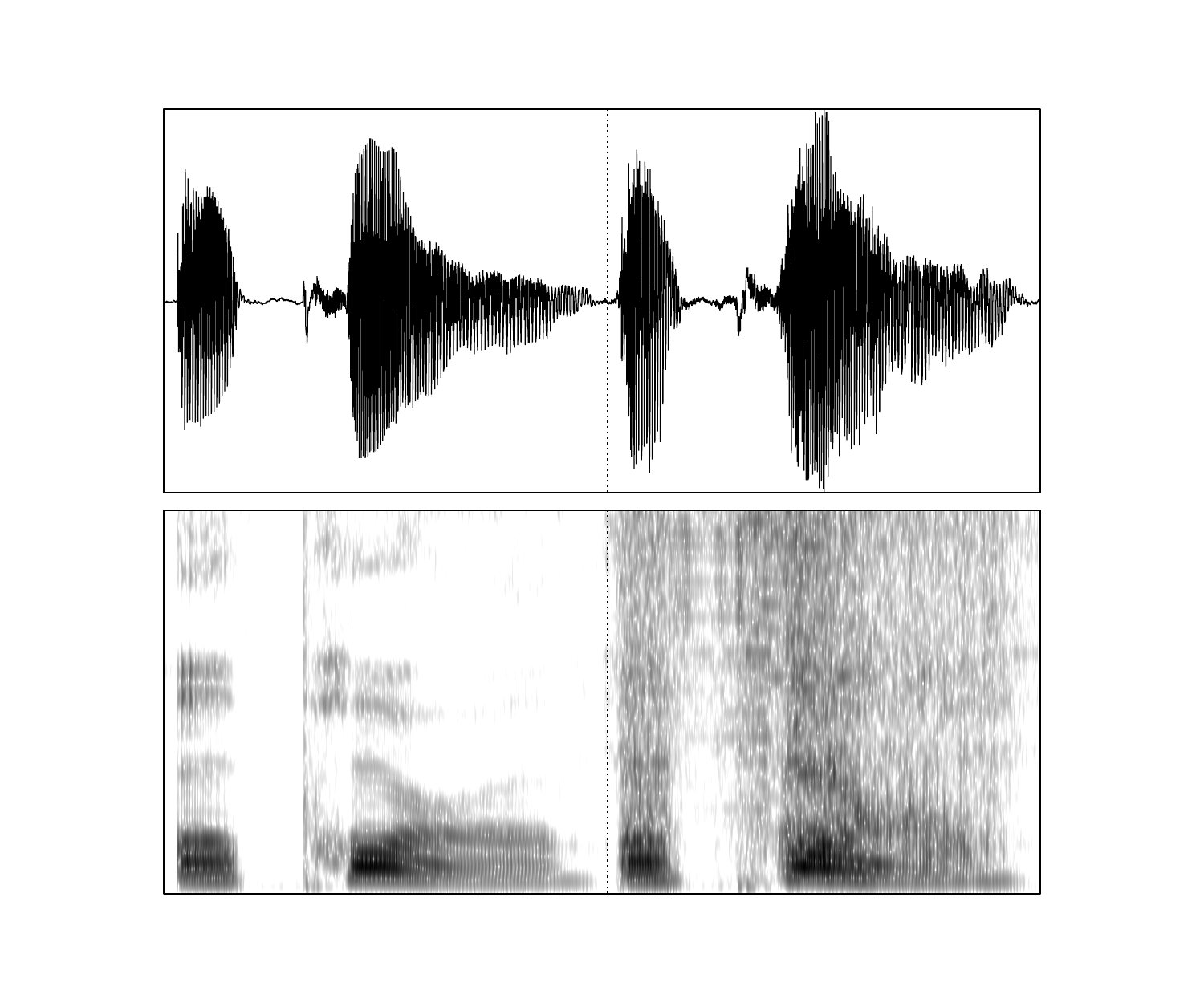}\\
\includegraphics[width=.5\textwidth]{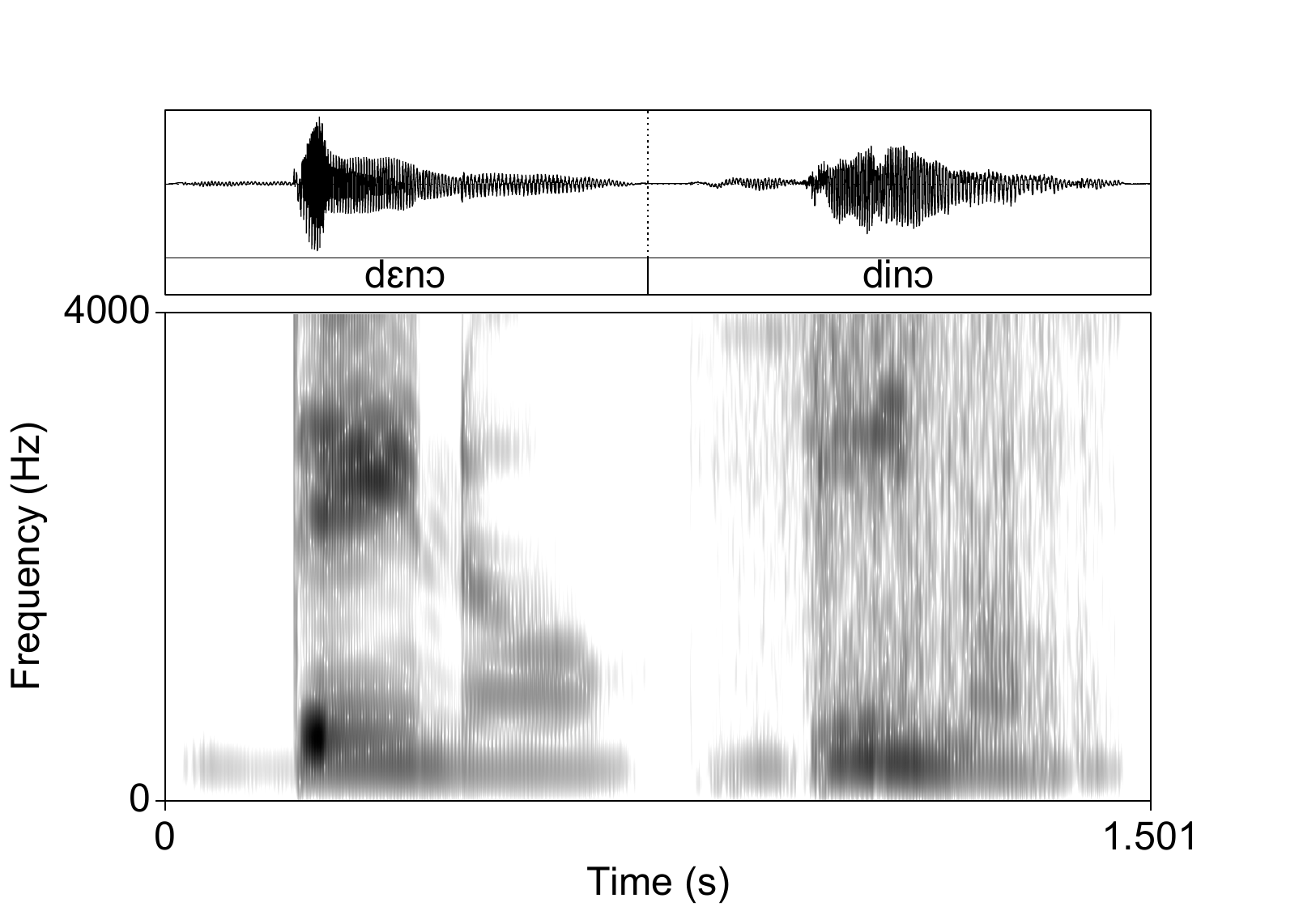}
\caption{\label{dino7453ACL2020}Waveforms and spectrograms (0--4,000 Hz) of \textbf{(top)} input sample (left) and generated sample (right) of [\textipa{O"p\super hORO}]; and \textbf{(bottom)} input sample [\textipa{"dEnO}] (left) and generated sample (right) [\textipa{"dinO}]. Both outputs are from models trained after 7453 steps.} 
\end{figure}

\begin{table}[H]\centering
\caption{\label{nonce1} IPA transcriptions and orthography of training data without consonantal changes for prefix VN-; C$_1$ is a sonorant. [\textipa{"lu\*ru}]	and	[\textipa{On"lu\*ru}] are missing from the computational experiment.}
\begin{tabular}{llllll}
\hline\hline
\multicolumn{6}{c}{\textbf{Fillers}}\\
\textbf{\#\underline{\hskip1em}}&\textbf{Harm.}&\textbf{Sg.}&\textbf{Pl.}&\multicolumn{2}{c}{\textbf{Orthography}}\\\hline
\multirow{4}{*}{[l]}&\multirow{2}{*}{[+fr]}&\textipa{"lEn} 	&	\textipa{En"lEn}&len&enlen	\\
&&\textipa{"linO}	&	\textipa{En"linO}&lino&enlino	\\
&\multirow{2}{*}{[$-$fr]}&\textipa{"lO\*r}	&	\textipa{On"lO\*r}&lor&onlor	\\
&&\textipa{"lu\*ru}	&	\textipa{On"lu\*ru}&luru&onluru	\\
\hline
\multirow{4}{*}{[r]}&\multirow{2}{*}{[+fr]}&\textipa{"\*rEl}	&	\textipa{En"\*rEl}	&rel&enrel\\
&&\textipa{"\*rinu}	&	\textipa{En"\*rinu}&rinu&enrinu	\\

&\multirow{2}{*}{[$-$fr]}&\textipa{"\*rAs}	&	\textipa{On"\*rAs}&ras&onras	\\
&&\textipa{"\*rOlO}	&	\textipa{On"\*rOlO}	&rolo&onrolo\\\hline

\multirow{4}{*}{[j]}&\multirow{2}{*}{[+fr]}&\textipa{"jim}	&	\textipa{En"jim}	&yim & enyim\\
&&\textipa{"jeni}	&	\textipa{En"jEni}&yeni&enyeni	\\
&\multirow{2}{*}{[$-$fr]}&\textipa{"jAm}	&	\textipa{On"jAm}&yam&onyam	\\
&&\textipa{"jAlu}	&	\textipa{On"jAlu}&yalu&onyalu	\\

\hline\hline
\end{tabular}
\end{table}

\begin{table}[H]\centering
\caption{\label{nonce1a} IPA transcriptions and orthography of training data without consonantal changes for prefix V-; C$_1$ is a sonorant.}

\begin{tabular}{llllll}
\hline\hline
\multicolumn{6}{c}{\textbf{Fillers}}\\
\textbf{\#\underline{\hskip1em}}&\textbf{Harm.}&\textbf{Sg.}&\textbf{Pl.}&\multicolumn{2}{c}{\textbf{Orthography}}\\\hline
\multirow{4}{*}{[l]}&\multirow{2}{*}{[+fr]}&\textipa{"lEm} 	&	\textipa{E"lEm}&lem&elem	\\
&&\textipa{"linO}	&	\textipa{E"linO}&lino&elino	\\
&\multirow{2}{*}{[$-$fr]}&\textipa{"lO\*r}	&	\textipa{O"lO\*r}&lor&olor	\\
&&\textipa{"lu\*ru}	&	\textipa{O"lu\*ru}&luru&oluru	\\\hline
\multirow{4}{*}{[r]}&\multirow{2}{*}{[+fr]}&\textipa{"\*rEl}	&	\textipa{E"\*rEl}	&rel&erel\\
&&\textipa{"\*rinu}	&	\textipa{E"\*rinu}&rinu&erinu	\\

&\multirow{2}{*}{[$-$fr]}&\textipa{"\*rAs}	&	\textipa{O"\*rAs}&ras&oras	\\
&&\textipa{"\*rOlO}	&	\textipa{O"\*rOlO}	&rolo&orolo\\\hline

\multirow{4}{*}{[j]}&\multirow{2}{*}{[+fr]}&\textipa{"jim}	&	\textipa{E"jim}	&yim & eyim\\
&&\textipa{"jeni}	&	\textipa{E"jEni}&yeni&eyeni	\\
&\multirow{2}{*}{[$-$fr]}&\textipa{"jAm}	&	\textipa{O"jAm}&yam&oyam	\\
&&\textipa{"jAlu}	&	\textipa{O"jAlu}&yalu&oyalu	\\

\hline\hline
\end{tabular}
\end{table}

\begin{table}[H]\centering
\caption{\label{nonce2} IPA transcriptions and orthography of training data  without consonantal changes for prefix VN-; C$_1$ is a voiceless obstruent.}\scalebox{.7}{
\begin{tabular}{lllllll}
\hline\hline
\multicolumn{7}{c}{\textbf{Voiceless}}\\
\textbf{Place}&\textbf{\#\underline{\hskip1em}}&\textbf{Harm.}&\textbf{Sg.}&\textbf{Pl.}&\multicolumn{2}{c}{\textbf{Orthography}}\\\hline

\multirow{7}{*}{Labial}&\multirow{3}{*}{[$-$cont]}	&	\multirow{2}{*}{[+fr]}&	\textipa{"p\super hin@}	&	\textipa{Em"p\super hin@}&pina&empina	\\
&	&	&\textipa{"p\super himi}&\textipa{Em"p\super himi}&	pimi	&	empimi	\\
&	&	[$-$fr]&	\textipa{"p\super hO\*rO}	&	\textipa{Om"p\super hO\*rO} &poro&omporo	\\\cline{2-7}

&\multirow{4}{*}{[$+$cont]}	&\multirow{2}{*}{[+fr]}	&\textipa{"fini}&\textipa{Em"fini}&fini	&	emfini	\\
&	&	&\textipa{"fim@}&\textipa{Em"fim@}&	fima	&	emfima	\\
&	&\multirow{2}{*}{[$-$fr]}	&\textipa{"fu\*r@}&\textipa{Om"fu\*r@}&	fura	&	omfura	\\
&	&	&\textipa{"fOlO}&\textipa{Om"fOlO}&	folo	&	omfolo	\\\hline

\multirow{7}{*}{Coronal}&\multirow{3}{*}{[$-$cont]}	&\multirow{2}{*}{[+fr]}	&\textipa{"t\super hElO}&\textipa{En"t\super hElO}&	telo	&	entelo	\\
&	&	&\textipa{"t\super hin@}&\textipa{En"t\super hin@}&	tina	&	entina	\\
&	&	[$-$fr]&	\textipa{"t\super hA\*ru}&\textipa{On"t\super hA\*ru}&taru	&	ontaru	\\\cline{2-7}

&\multirow{4}{*}{[$+$cont]}&	\multirow{2}{*}{[+fr]}&	\textipa{"sEnO}&\textipa{En"sEnO}&seno	&	enseno	\\
&	&	&\textipa{"sil@}	&\textipa{En"sil@}	&sila	&	ensila	\\
&	&	\multirow{2}{*}{[$-$fr]}&	\textipa{"sO\*rO}&\textipa{On"sO\*rO}&soro	&	onsoro	\\
&	&	&	\textipa{"sAnu}&\textipa{On"sAnu}&sanu	&	onsanu	\\

\hline\hline
\end{tabular}}
\end{table}

\begin{table}[H]\centering
\caption{\label{nonce2a} IPA transcriptions and orthography of training data  without consonantal changes for prefix V-; C$_1$ is a voiceless obstruent.}\scalebox{.7}{
\begin{tabular}{lllllll}
\hline\hline
\multicolumn{7}{c}{\textbf{Voiceless}}\\
\textbf{Place}&\textbf{\#\underline{\hskip1em}}&\textbf{Harm.}&\textbf{Sg.}&\textbf{Pl.}&\multicolumn{2}{c}{\textbf{Orthography}}\\\hline

\multirow{7}{*}{Labial}&\multirow{3}{*}{[$-$cont]}	&	\multirow{2}{*}{[+fr]}&	\textipa{"p\super hin@}	&	\textipa{E"p\super hin@}&pina&epina	\\
&	&	&\textipa{"p\super himi}&\textipa{E"p\super himi}&	pimi	&	epimi	\\
&	&	\multirow{2}{*}{[$-$fr]}&	\textipa{"p\super hO\*rO}	&	\textipa{O"p\super hO\*rO} &poro&oporo	\\
&	&	&	\textipa{"p\super hO\*mO}	&	\textipa{O"p\super hO\*mO} &pomo&opomo	\\
\cline{2-7}

&\multirow{4}{*}{[$+$cont]}	&\multirow{2}{*}{[+fr]}	&\textipa{"fini}&\textipa{E"fini}&fini	&	efini	\\
&	&	&\textipa{"fim@}&\textipa{E"fim@}&	fima	&	efima	\\
&	&\multirow{2}{*}{[$-$fr]}	&\textipa{"fu\*r@}&\textipa{O"fu\*r@}&	fura	&	ofura	\\
&	&	&\textipa{"fOlO}&\textipa{O"fOlO}&	folo	&	ofolo	\\\hline

\multirow{7}{*}{Coronal}&\multirow{3}{*}{[$-$cont]}	&\multirow{2}{*}{[+fr]}	&\textipa{"t\super hElO}&\textipa{E"t\super hElO}&	telo	&	etelo	\\
&	&	&\textipa{"t\super hin@}&\textipa{E"t\super hin@}&	tina	&	etina	\\
&	&	\multirow{2}{*}{[-fr]}&	\textipa{"t\super hA\*ru}&\textipa{O"t\super hA\*ru}&taru	&	otaru	\\
&	&	&	\textipa{"t\super hOmO}&\textipa{O"t\super hOmO}&tomo	&	otomo	\\
\cline{2-7}

&\multirow{4}{*}{[$+$cont]}&	\multirow{2}{*}{[+fr]}&	\textipa{"sEnO}&\textipa{E"sEnO}&seno	&	eseno	\\
&	&	&\textipa{"sil@}	&\textipa{E"sil@}	&sila	&	esila	\\
&	&	\multirow{2}{*}{[$-$fr]}&	\textipa{"sO\*rO}&\textipa{O"sO\*rO}&soro	&	osoro	\\
&	&	&	\textipa{"sAnu}&\textipa{O"sAnu}&sanu	&	osanu	\\

\hline\hline
\end{tabular}}
\end{table}

\begin{table}[H]\centering
\caption{\label{nonce3} IPA transcriptions and orthography of training data with consonantal changes for prefix VN-.}\scalebox{.7}{
\begin{tabular}{lllllll}
\hline\hline
\multicolumn{7}{c}{\textbf{Voiced}}\\
\textbf{Place}&\textbf{\#\underline{\hskip1em}}&\textbf{Harm.}&\textbf{Sg.}&\textbf{Pl.}&\multicolumn{2}{c}{\textbf{Orthography}}\\\hline

\multirow{16}{*}{Labial}&\multirow{8}{*}{[$-$cont]}	&\multirow{4}{*}{[+fr]}	&\textipa{"bil@}&\textipa{Em"p\super hil@}&	bila	&	empila	\\
&	&	&\textipa{"be\*r@}&\textipa{Em"p\super he\*r@}&	bera	&	empera	\\
&	&	&\textipa{"bilO}&\textipa{Em"p\super hilO}&	bilo	&	empilo	\\
&	&	&	\textipa{"bEm@}&\textipa{Em"p\super hEm@}&bema	&	empema	\\\cline{3-7}
&	&	\multirow{4}{*}{[$-$fr]}&\textipa{"bul@}&\textipa{Om"p\super hul@}&	bula	&	ompula	\\
&	&	&	\textipa{"bAlu}&\textipa{Om"p\super hAlu}&balu	&	ompalu	\\
&	&	&	\textipa{"bO\*r@}&\textipa{Om"pO\*r@}&bora	&	ompora	\\
&	&	&\textipa{"bunE}&\textipa{Om"punE}&	bune	&	ompune	\\\cline{2-7}

&	\multirow{8}{*}{[$+$cont]}&	\multirow{4}{*}{[+fr]}&\textipa{"vil@}&\textipa{Em"p\super hil@}&	vila	&	empila	\\
&	&	&\textipa{"vEmO}&\textipa{Em"p\super hEmO}&	vemo	&	empemo	\\
&	&	&\textipa{"vi\*r@}&\textipa{Em"p\super hi\*r@}&	vira	&	empira	\\
&	&	&	\textipa{"vEl@}&\textipa{Em"p\super hEl@}&vela	&	empela	\\\cline{3-7}
&	&	\multirow{4}{*}{[$-$fr]}&\textipa{"vulO}&\textipa{Om"p\super hulO}&	vulo	&	ompulo	\\
&	&	&\textipa{"vA\*ru}&\textipa{Om"p\super hA\*ru}&	varu	&	omparu	\\
&	&	&\textipa{"vOn@}&\textipa{Om"p\super hOn@}&	vona	&	ompona	\\
&	&	&\textipa{"vulE}&\textipa{Om"p\super hulE}&	vule	&	ompule	\\\hline

\multirow{16}{*}{Coronal}&		\multirow{8}{*}{[$-$cont]}&	\multirow{4}{*}{[+fr]}&	\textipa{"dilO}&\textipa{En"t\super hilO}&dilo	&	entilo	\\
&	&	&\textipa{"di\*ri}&\textipa{En"t\super hi\*ri}&	diri	&	entiri	\\
&	&	&\textipa{"dElO}&\textipa{En"t\super hElO}&	delo	&	entelo	\\
&	&	&\textipa{"dEm@}&\textipa{En"t\super hEm@}&	dema	&	entema	\\\cline{3-7}
&	&\multirow{4}{*}{[$-$fr]}	&\textipa{"dulE}&\textipa{On"t\super hulE}&	dule	&	ontule	\\
&	&	&\textipa{"dO\*ru}&\textipa{On"t\super hO\*ru}&	doru	&	ontoru	\\
&	&	&\textipa{"dAlE}&\textipa{On"t\super hAlE}&	dale	&	ontale	\\
&	&	&\textipa{"dun@}&\textipa{On"t\super hun@}&	duna	&	ontuna	\\\cline{2-7}

&	\multirow{8}{*}{[$+$cont]}&	\multirow{4}{*}{[+fr]}&\textipa{"zil@}&\textipa{En"t\super hil@}&	zila	&	entila	\\
&	&	&\textipa{"zi\*r@}&\textipa{En"t\super hi\*r@}&	zira	&	entira	\\
&	&	&	\textipa{"zEmO}	&\textipa{En"t\super hEmO}&zemo&	entemo	\\
&	&	&\textipa{"zEni}&\textipa{En"t\super hEni}&	zeni	&	enteni	\\\cline{3-7}
&	&	\multirow{4}{*}{[$-$fr]}&	\textipa{"zulO}&\textipa{On"t\super hulO}&zulo	&	ontulo	\\
&	&	&\textipa{"zA\*ru}&\textipa{On"t\super hA\*ru}&	zaru	&	ontaru	\\
&	&	&\textipa{"zOlE}&\textipa{On"t\super hOlE}&	zole	&	ontole	\\
&	&	&\textipa{"zunE}&\textipa{On"t\super hunE}&	zune	&	ontune	\\

\hline\hline
\end{tabular}}
\end{table}

\begin{table}[H]\centering
\caption{\label{nonce3a} IPA transcriptions of training data with consonantal changes for prefix V-.}\scalebox{.7}{
\begin{tabular}{lllllll}
\hline\hline
\multicolumn{7}{c}{\textbf{Voiced}}\\
\textbf{Place}&\textbf{\#\underline{\hskip1em}}&\textbf{Harm.}&\textbf{Sg.}&\textbf{Pl.}&\multicolumn{2}{c}{\textbf{Orthography}}\\\hline

\multirow{16}{*}{Labial}&\multirow{8}{*}{[$-$cont]}	&\multirow{4}{*}{[+fr]}	&\textipa{"bElO}&\textipa{E"p\super hElO}&	belo	&	epelo	\\
&	&	&\textipa{"bel@}&\textipa{E"p\super hel@}&	bela	&	epela	\\
&	&	&\textipa{"bi\*r@}&\textipa{E"p\super hi\*r@}&	bira	&	epira	\\
&	&	&	\textipa{"bim@}&\textipa{E"p\super him@}&bima	&	epima	\\\cline{3-7}
&	&	\multirow{4}{*}{[$-$fr]}&\textipa{"bulE}&\textipa{O"p\super hulE}&	bule	&	opule	\\
&	&	&	\textipa{"bA\*ru}&\textipa{O"p\super hA\*ru}&baru	&	oparu	\\
&	&	&	\textipa{"bulO}&\textipa{O"pulO}&bulo	&	opulo	\\
&	&	&\textipa{"bOn@}&\textipa{O"pOn@}&	bona	&	opona	\\\cline{2-7}

&	\multirow{8}{*}{[$+$cont]}&	\multirow{4}{*}{[+fr]}&\textipa{"bilO}&\textipa{E"filO}&	bilo	&	efilo	\\
&	&	&\textipa{"bEm@}&\textipa{E"fEm@}&	bema	&	efema	\\
&	&	&\textipa{"bil@}&\textipa{E"fil@}&	bila	&	efila	\\
&	&	&	\textipa{"bE\*rO}&\textipa{E"fE\*rO}&bero	&	efero	\\\cline{3-7}
&	&	\multirow{4}{*}{[$-$fr]}&\textipa{"bul@}&\textipa{O"ful@}&	bula	&	ofula	\\
&	&	&\textipa{"bAlu}&\textipa{O"fAlu}&	balu	&	ofalu	\\
&	&	&\textipa{"bO\*r@}&\textipa{O"fO\*r@}&	bora	&	ofora	\\
&	&	&\textipa{"bunE}&\textipa{O"funE}&	bune	&	ofune	\\\hline

\multirow{16}{*}{Coronal}&		\multirow{8}{*}{[$-$cont]}&	\multirow{4}{*}{[+fr]}&	\textipa{"dil@}&\textipa{E"t\super hil@}&dila	&	etila	\\
&	&	&\textipa{"di\*ru}&\textipa{E"t\super hi\*ru}&	diru	&	etiru	\\
&	&	&\textipa{"dEni}&\textipa{E"t\super hEni}&	deni	&	eteni	\\
&	&	&\textipa{"dEm@}&\textipa{E"t\super hEm@}&	dema	&	etema	\\\cline{3-7}
&	&\multirow{4}{*}{[$-$fr]}	&\textipa{"dulO}&\textipa{O"t\super hulO}&	dulo	&	otulo	\\
&	&	&\textipa{"dA\*ru}&\textipa{O"t\super hA\*ru}&	daru	&	otaru	\\
&	&	&\textipa{"dOlE}&\textipa{O"t\super hOlE}&	dole	&	otole	\\
&	&	&\textipa{"dunE}&\textipa{O"t\super hunE}&	dune	&	otune	\\\cline{2-7}

&	\multirow{8}{*}{[$+$cont]}&	\multirow{4}{*}{[+fr]}&\textipa{"dilu}&\textipa{E"silu}&	dilu	&	esilu	\\
&	&	&\textipa{"di\*ri}&\textipa{E"si\*ri}&	diri	&	esiri	\\
&	&	&	\textipa{"dEmE}	&\textipa{E"sEmE}&deme&	eseme	\\
&	&	&\textipa{"dEnO}&\textipa{E"sEnO}&	deno	&	eseno	\\\cline{3-7}
&	&	\multirow{4}{*}{[$-$fr]}&	\textipa{"dulE}&\textipa{O"sulE}&dule	&	osule	\\
&	&	&\textipa{"dO\*ru}&\textipa{O"sO\*ru}&	doru	&	osoru	\\
&	&	&\textipa{"dAl@}&\textipa{O"sAl@}&	dala &	osala	\\
&	&	&\textipa{"dun@}&\textipa{O"sun@}&	duna&	osuna	\\

\hline\hline
\end{tabular}}
\end{table}

\begin{table}[H]\centering
\caption{\label{test3a} IPA transcriptions and orthography of training data without the prefixed forms.}
\begin{tabular}{llllllllllll}
\hline\hline
\textipa{"bA\*r@}&bara&\textipa{"vA\*r@}&vara&\textipa{"dAmi}&dami&\textipa{"zAmi}&zami&\textipa{"lEni} (2$\times$)&leni&\textipa{"\*rEm@} ($2\times$)&rema\\
\textipa{"bAj@} (2$\times$)&baja&\textipa{"vAj@}&vaya&\textipa{"dAwE}&dawe&\textipa{"zAwO}&zawo&\textipa{"li\*rO} (2$\times$)&liro&\textipa{"\*ru\*rO} ($2\times$)&ruro\\
\textipa{"bEnE}&bene&\textipa{"vEnE}&vene&\textipa{"dAwO}&dawo&\textipa{"zElE}&zele&\textipa{"lOna} (2$\times$)&lona&& \\
\textipa{"bEjO} (2$\times$)&beyo&\textipa{"vEjo}&vejo&\textipa{"dElE}&dele&\textipa{"ziwO}&ziwo&\textipa{"lOnu} (2$\times$)&lonu &&\\
\textipa{"bijE} &biye&&&\textipa{"dEwE}&dewe&\textipa{}&&\textipa{}&&&\\
\textipa{"bujE} &buye&&&\textipa{"diwO} (2$\times$)&diwo &&\textipa{}&&&\\
\textipa{} &\textipa{}&&&\textipa{"dOw@}&\textipa{dowa}&&&&&&\\

\hline\hline
\end{tabular}
\end{table}

\begin{table}[ht]
\centering
\begin{tabular}{rrrrr}
\hline   \hline
 & Estimate & Std. Error & z value & Pr($>$$|$z$|$) \\ 
  \hline
(Intercept) = mean& 1.34 & 0.19 & 7.20 & 0.0000 \\ 
mean vs.~back  & 0.30 & 0.19 & 1.64 & 0.1016 \\ 
mean vs.~V- & 0.05 & 0.19 & 0.29 & 0.7710 \\ 
  Frontness:Prefix & -0.05 & 0.19 & -0.29 & 0.7710 \\ 
\hline   \hline
\end{tabular}
\caption{\label{genLinModel}Linear  logistic regression estimates with harmonious responses of the Generator network as successes and vowel \textsc{frontnesss} (with two sum-coded levels, front and back) and \textsc{prefix} identity (with two sum-coded levels, V- and VN-) as the independent variables with their interaction.}
\end{table}

\begin{table}[ht]
\centering
\begin{tabular}{rrrrr}
  \hline\hline
 & Estimate & Std. Error & z value & Pr($>$$|$z$|$) \\ 
  \hline
(Intercept) & 0.56 & 0.11 & 5.01 & 0.0000 \\ 
  frontness1 & 0.08 & 0.11 & 0.75 & 0.4549 \\ 
  prefix1 & 0.04 & 0.06 & 0.72 & 0.4738 \\ 
  frontness1:prefix1 & 0.09 & 0.05 & 1.86 & 0.0623 \\ 
   \hline\hline
\end{tabular}

\caption{\label{subjectsExperiment}Linear mixed effects logistic regression estimates with harmonious responses of human subjects in the behavioral experiment as successes and vowel \textsc{frontnesss} (with two sum-coded levels, front and back) and \textsc{prefix} identity (VN- vs.~V-, sum-coded) as the independent variables with their interaction.}
\end{table}

\begin{table}[ht]
\centering
\begin{tabular}{lrrrr}
   \hline   \hline
A. parametric coefficients & Estimate & Std. Error & t-value & p-value \\ 
  (Intercept) & 1.3840 & 0.2738 & 5.0543 & $<$ 0.0001 \\ 
   \hline
B. smooth terms & edf & Ref.df & F-value & p-value \\ 
  s(traj) & 1.0000 & 1.0000 & 60.9063 & $<$ 0.0001 \\ 
  s(traj,latent) & 90.2991 & 489.0000 & 223.4940 & $<$ 0.0001 \\ 
   \hline   \hline
\end{tabular}
\caption{\label{interpolation1}Estimates of a generalized additive mixed effects logistic regression model with the front vs.~back triggering vowel (V$_2$) value (front = success; back = failure)  as the dependent variable and a thin-plate smooth for values of $z_{17}$ as the independent variable (with random smooths for each of the 60 generated sets).} 
\end{table}

\begin{table}[ht]
\centering
\begin{tabular}{lrrrr}
   \hline   \hline
A. parametric coefficients & Estimate & Std. Error & t-value & p-value \\ 
  (Intercept) = back& 1.4305 & 0.3379 & 4.2333 & $<$ 0.0001 \\ 
  frontness = back vs.~front  & -0.0391 & 0.3536 & -0.1106 & 0.9119 \\ 
   \hline
B. smooth terms & edf & Ref.df & F-value & p-value \\ 
  s(traj):frontness = back & 1.0000 & 1.0000 & 0.7032 & 0.4017 \\ 
  s(traj):frontness = front& 2.5933 & 3.1648 & 6.9795 & 0.0813 \\ 
  s(traj,latent) & 94.3537 & 489.0000 & 210.2151 & $<$ 0.0001 \\ 
   \hline   \hline
\end{tabular}
\caption{\label{interpolation2}Estimates of a generalized additive mixed effects logistic regression model with harmonious (success) and disharmonious (failure) outcome as the dependent variable, vowel \textsc{frontness} as a parametric predictor, and thin-plate smooths for the two levels of frontness (front vs.~back, treatment-coded with back as the reference level) across the values of $z_{17}$, and random smooths for each of the 60 set of generated outputs. } 
\end{table}

\end{document}